\documentclass[pdflatex,sn-mathphys-num]{sn-jnl}

\usepackage[english]{babel}

\usepackage{float}
\usepackage{graphicx}
\usepackage{subfig}
\usepackage{overpic}

\usepackage{amsthm}
\usepackage{amsmath,amssymb,amsfonts}
\usepackage{bm}
\usepackage{listings}

\usepackage[figuresright]{rotating}
\usepackage{algorithmic}
\usepackage[ruled,vlined,linesnumbered]{algorithm2e}

\usepackage{array}
\usepackage{booktabs}
\usepackage{enumerate}
\usepackage{multirow}

\usepackage{url}
\usepackage{color}

\theoremstyle{thmstyleone}%
\theoremstyle{thmstyletwo}%

\theoremstyle{thmstylethree}%

\begin{document}

\title[Article Title]{Design and Control of the ``QuadBoat": A Quadruped Surface Vehicle for Drowning Rescue}


\author[1,2]{\fnm{Lianxin} \sur{Zhang}}\email{lianxinzhang@link.cuhk.edu.cn}

\author[3]{\fnm{Yihan} \sur{Huang}}\email{yhuang101@stevens.edu}

\author*[1,2]{\fnm{Huihuan} \sur{Qian}}\email{hhqian@cuhk.edu.cn}

\affil*[1]{\orgname{Shenzhen Institute of Artificial Intelligence and Robotics for Society (AIRS)}, \orgaddress{\city{Shenzhen}, \postcode{518129}, \state{Guangdong}, \country{China}}}

\affil[2]{\orgdiv{School of Science and Engineering}, \orgname{The Chinese University of Hong Kong, Shenzhen (CUHK-Shenzhen)}, \orgaddress{\city{Shenzhen}, \postcode{518172}, \state{Guangdong}, \country{China}}}

\affil[3]{\orgdiv{Department of Electrical and Computer Engineering}, \orgname{Stevens Institute of Technology}, \orgaddress{\city{Hoboken}, \postcode{07030}, \state{New Jersey}, \country{United States}}}


\abstract{
	Prompt extraction of victims from water is crucial in water surface rescue missions. However, previous research on rescue robots has seldom addressed this issue.
	This paper presents QuadBoat, a bio-inspired unmanned surface vehicle (USV) designed to track and retrieve victims from water.
	QuadBoat features a quadrupedal robot configuration, enabling it with highly adaptable and agile maneuverability through its actively adjustable posture. 
	Employing an inverse kinematics-based controller and cascaded model predictive control (MPC)-PID controller for overall movement, QuadBoat can accurately track and retrieve objects on the water surface. 
	Maneuverability demonstrations validate QuadBoat's high agility, while a series of tracking experiments, including leg action tracking and trajectory tracking, confirm its high motion accuracy and system mobility.
	Finally, visual-based tracking and object pickup experiments further verify QuadBoat's target tracking capabilities and its effectiveness in executing rescues, both indoors and outdoors.
}




\maketitle

\section{Introduction}\label{Sect:intro}
Over the last decade, drowning has emerged as a significant global public health concern, contributing to an annual toll of over 0.2 million fatalities \cite{Organization2023}.
Consequently, this matter has garnered attention from both society and the academic community.
Drowning incidents can occur in various circumstances, such as pools, rivers, lakes, or oceans.
When it happens, the victim may face temporary or permanent disability, or even death, within 5-10 minutes. 
Thus, in the rescue, it is of vital importance to promptly and effectively localize the drowning victims and pull them out of the water. 
Robotic systems, utilizing various sensor information, have shown tremendous potential in search and rescue (SAR), and can ensure the safety of rescuers.
Among various rescue robots, such as unmanned aerial vehicles (UAVs) \cite{Albanese2021,Lyu2023}, amphibious robots \cite{Kim2023}, and underwater robots \cite{Ribas2015,Bonfitto2018,Wu2021}, unmanned surface vehicles (USVs) have advantages in terms of high payload capacity, strong maneuverability, efficiency, and reliability.
Therefore, in this paper, we are interested in studying innovative USVs for drowning rescue.

\begin{figure} [tbp] 
	\centering
	\includegraphics[width=0.7\linewidth]{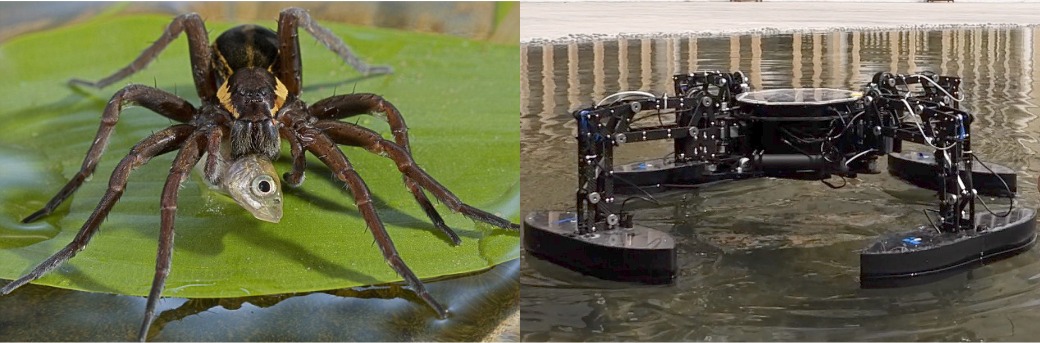}
	\caption{Inspired by fisher spiders in nature, the QuadBoat can maneuver on the water surface and collect floating objects. }
	\label{fig:design_idea}
\end{figure}

The rescue process can be divided into three steps: searching and capturing the victim, extracting them from the water, and transporting them to a safe place. 
Here, we focus on the latter two capabilities of the robot.

\subsection{Related Work}


In nature, aquatic or semiaquatic animals, such as the fisher spider and water strider, pick up objects or prey for purposes like nesting or predation  \cite{Shimizu1992,Nyffeler2014}, as depicted in Fig. \ref{fig:design_idea}.
Inspired by their locomotion and payload capacity, numerous biomimetic robots have been developed to explore the design of water surface robotic systems.
In existing studies, robots of different sizes utilize unique mechanisms for movement and flotation on the water \cite{Kim2024}.
For instance, in a small scale (body length $< 10^{-1}$ m),  robots rely on surface tension-dominant drag at the air-liquid interface and leverage wire legs coated with hydrophobic material to balance their weight \cite{Song2007,Yan2020,Kim2022}.
Based on this design, such robots are endowed with capability of walking or even jumping on water surface \cite{Koh2015,Gwon2023}.
Comparatively, to improve the hydrodynamic supporting forces, robots in larger scale (body length $> 10^{-1}$ m) employ large footpads or air balls at the end of legs to obtain buoyancy \cite{Yang2016,Sun2018,Chen2018,Zhang2022}.
However, the water strider-inspired design utilizing surface tension is not applicable at macroscopic scales. 
Additionally, the aforementioned robotic systems have limited payload capacity, rendering them incapable of undertaking the rescue missions.

Drawing idea from the above biomimetic robots, rescue USVs can adopt a multi-hull design with supporting legs.
Moreover, according to the evaluation of heavy lift ships in \cite{Meng2022,Fang2023}, the multi-vessel float-over technology has more capacity and stability than the single-vessel one.
Over the past few years, numerous proposals for surface robots have been put forward regarding the SAR operations \cite{Jorge2019}.
For example, the Emergency Integrated Lifesaving Lanyards (EMILY), invented in 2010, is the first robust teleoperated USV for drowning rescue, which now also conduct tasks including gateway communications, surveillance, and infrastructure inspection \cite{Schofield2018,Wilde2018,Smith2022}.
The SWIFT USV is a low-cost small USV designed to carry support for humans in coastal waters or surf zones \cite{Matias2015}. 
It can move at a high speed (12 knots) with a light-weight hull and a directional water-jet propeller.
Meanwhile, USVs become increasingly prominent in exploration and search owing to their advantages such as high efficiency, cost-effectiveness,and swift deployment \cite{Li2023}.
The ROAZ catamaran can detect and tracking the victim using infrared (IR) and color cameras by coupling horizon detection with Hough transform, and rapidly get to the victim carrying and deploying survival capsules \cite{Matos2016}.
To detect the underwater scene, the PRIME catamaran is equipped with side?scan imaging sonars to acquire data and communicate this to an on-shore computer in real-time, which helps police to detect human bodies in shallow waters \cite{Rymansaib2023}. 
The collaborative systems of USVs and UAVs, leveraging their respective advantages, have also garnered significant attention for study \cite{Queralta2020}.
In these systems, UAVs garther and transmit real-time videos or images invisible from the ground for the localization and navigation of USVs \cite{Wang2023}.
However, these studies have predominantly focused on environmental exploration, target detection, and USV platform design, overlooking how robots can autonomously perform rescue actions such as extraction and transportation.

The rescue environment is often complex and unpredictable, demanding high maneuverability from the USV. 
In this paper, we prefer to adopt omnidirectional USVs, which encompass mono-hull and multi-hull types.
The mono-hull type comes in two configurations: with 3 propellers  \cite{Xue2023} and 4 propellers  \cite{Paulos2015,Wang2023a}. 
Research has shown that the 4-propeller configuration achieves higher propulsion efficiency \cite{Zhang2023}.
On the other side, the multi-hull type features a symmetric body design for payload expandability and damage redundancy \cite{Nad2015,Tao2018,Groves2020}, with some robots capable of shape transformation to enchance movement efficiency \cite{Mukai2021} and meet stability requirements \cite{Zhang2022a}.
%

\subsection{Contributions}
On these grounds, inspired by the hunting behavior of the fisher spider, we propose the QuadBoat, a novel omnidirectional USV designed for drowning rescue. The QuadBoat is equipped with four outrigger hulls installed at the end of its legs. 
With this quadrupedal structure, it is capable of standing on the water surface and extracting victims from the water. 
The QuadBoat is designed as a quad-maran with a symmetrical structure.
Its full actuation and transformation capabilities endow the QuadBoat with high maneuverability and adaptability in complex rescue environments.
It offers both fully-actuated and underactuated modes, providing precise control and high-speed movement, respectively.

The main contributions of this work are as follows:
\begin{enumerate} [1)]
	\item We introduce a biomimetic omnidirectional USV, the QuadBoat, designed specifically for water rescue missions. Inspired by the hunting behavior of fisher spiders, this robot can extract victims from the water and transport them to safety.
	\item We propose models for both leg action and overall movement of the QuadBoat, and we develop and experimentally validate a joint controller and a cascade model predictive control (MPC)-PID movement controller.
	\item We implement a functional prototype and validate its efficacy in water rescue through a vision-guided object extraction experiment, incorporating target tracking, extraction, and transportation.
\end{enumerate}

%

The remainder of this paper is organized as follows. 
In Section \ref{sect:system}, all hardware components of the QuadBoat are described in detail. 
Section \ref{sect:model} presents the kinematic model of the legs and the propulsion model for the QuadBoat.
Based on the established model, the PID controller for leg action and the MPC controller for movement are derived in Section \ref{sect:control}.
The demonstration tests and tracking experiments including leg actions, overall movement, and visual-guided tracking are illustrated and discussed in Section \ref{sect:experiment}.
Finally, Section \ref{sect:conclusion} concludes the paper.

\section{Hardware Design}
\label{sect:system}
\begin{figure} [htpb]
	\centering
	\subfloat[]{
		\centering
		\includegraphics[width=0.5\linewidth]{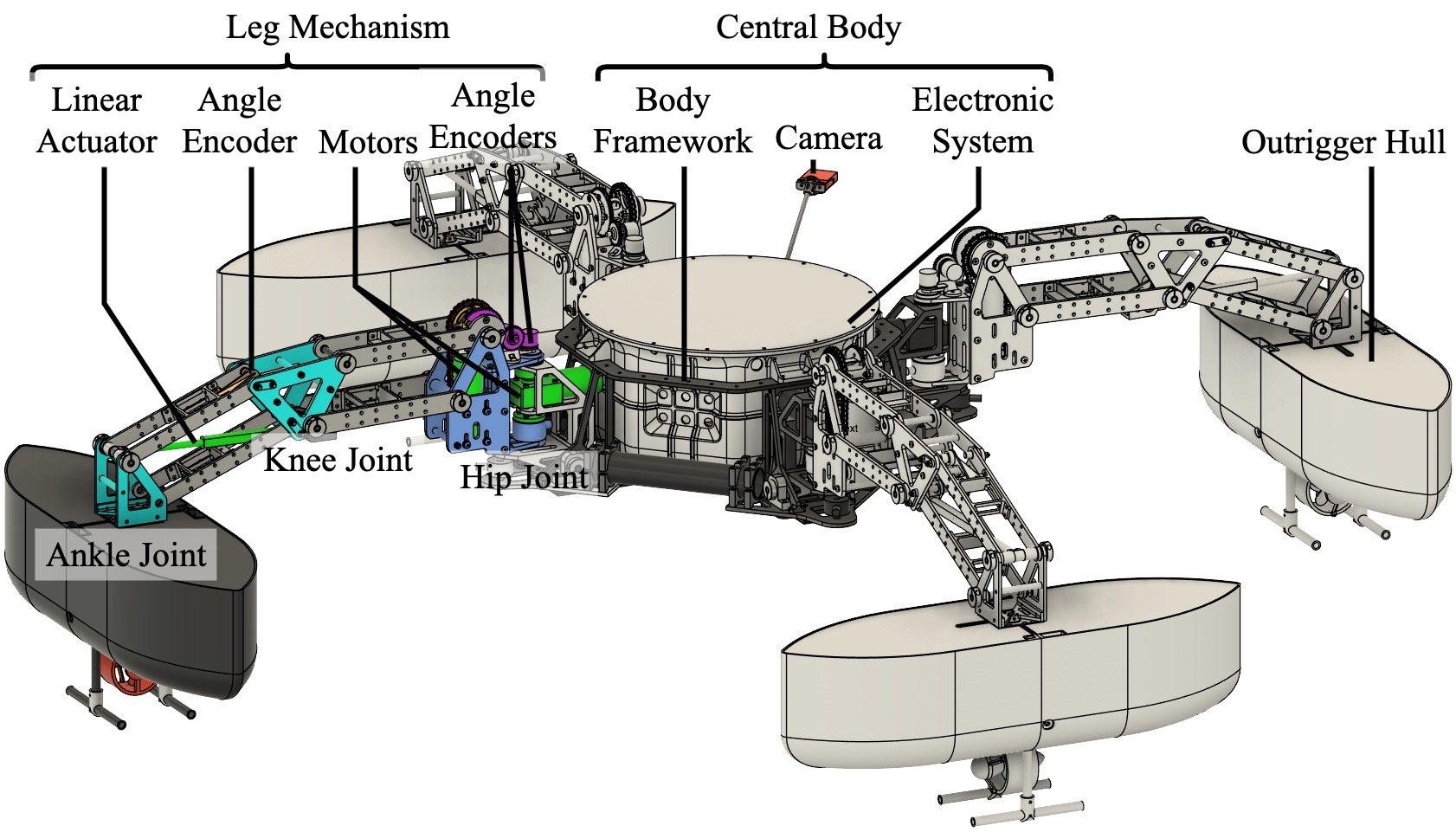}
	}
	\subfloat[]{
		\centering
		\includegraphics[width=0.5\linewidth]{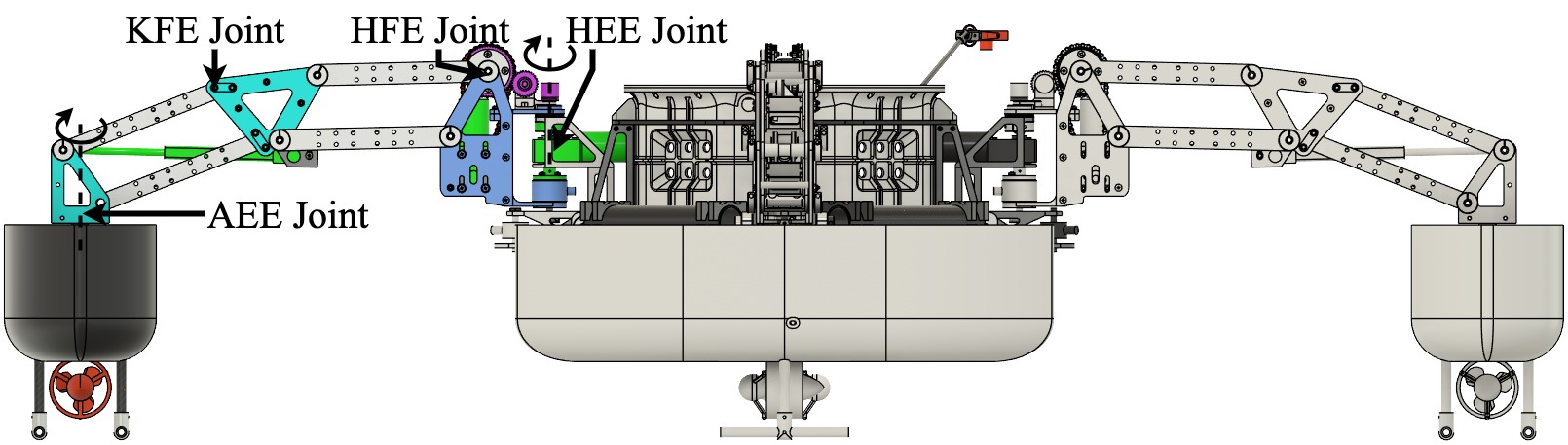}
	}
	\caption{(a) Main components and (b) front view of the QuadBoat.}
	\label{fig:body}
\end{figure}

The design goal is to create a robot that seamlessly integrates the ability to maintain balance while collecting floating objects or maneuvering on the water surface.
Fig. \ref{fig:body} (a) shows the main components of the QuadBoat design, including the mechanical structure and the electronic system, each of which is detailedly described in this section.

\subsection{Mechanical Design}

The QuadBoat features four outrigger hulls, each linked to a central body via a leg. 
Its overall dimensions vary within a width range of 1.188$\sim$1.85 m and a height range of 0.425$\sim$0.643 m.
The outrigger hulls incorporate streamlined buoyant structures with thrusters at the bottom, dedicated to providing buoyancy and thrust, respectively.
The central body predominantly accommodates electronic components, including battery packs, computing modules, motor drivers, etc. 
With four degrees of freedom (DoFs) per leg, the robot can lift, extend outward, and twist, allowing it to dynamically adapt to its environment.

%

\subsubsection{Legs}

%
The four leg mechanisms are all double-layered, two-stage serial parallelogram structures, symmetrically distributed around the central body.  
This structure, widely used in planetary rovers \cite{Cordes2018,He2022}, offers large workspace with few active DoFs. 
As shown in Fig. \ref{fig:body} (b), each leg comprises a hip joint with two DoFs and a knee joint with one DoF, namely the hip endo/exo-rotation (HEE) joint, hip flexion/ extension (HFE) joint, and knee flexion/extension(KFE) joint.
The three motors of the leg, including two DC motors with worm gear reducers of 670:1 ratio and one linear actuator, independently control the three DoFs.
The linear actuator can exert a maximum thrust of 60 N or a maximum speed of $0.012 \text{m/s}$, and the DC motors can generate a maximum torque of $9.81 \text{N} \cdot \text{m}$ or a maximum rotary speed of 12 rpm, ensuring the robot's capability to perform diverse movements.
Three encoders, installed at the joints, provide angle feedback for precise positioning of the leg endpoint in three dimensions relative to the body.
Thanks to this design, each QuadBoat leg can extend horizontally up to 0.332 m, collectively lifting the robot by a maximum of 0.435 m.


\subsubsection{Outrigger Hulls}

\begin{figure} [htpb]
	\centering
	\includegraphics[width=0.6\linewidth]{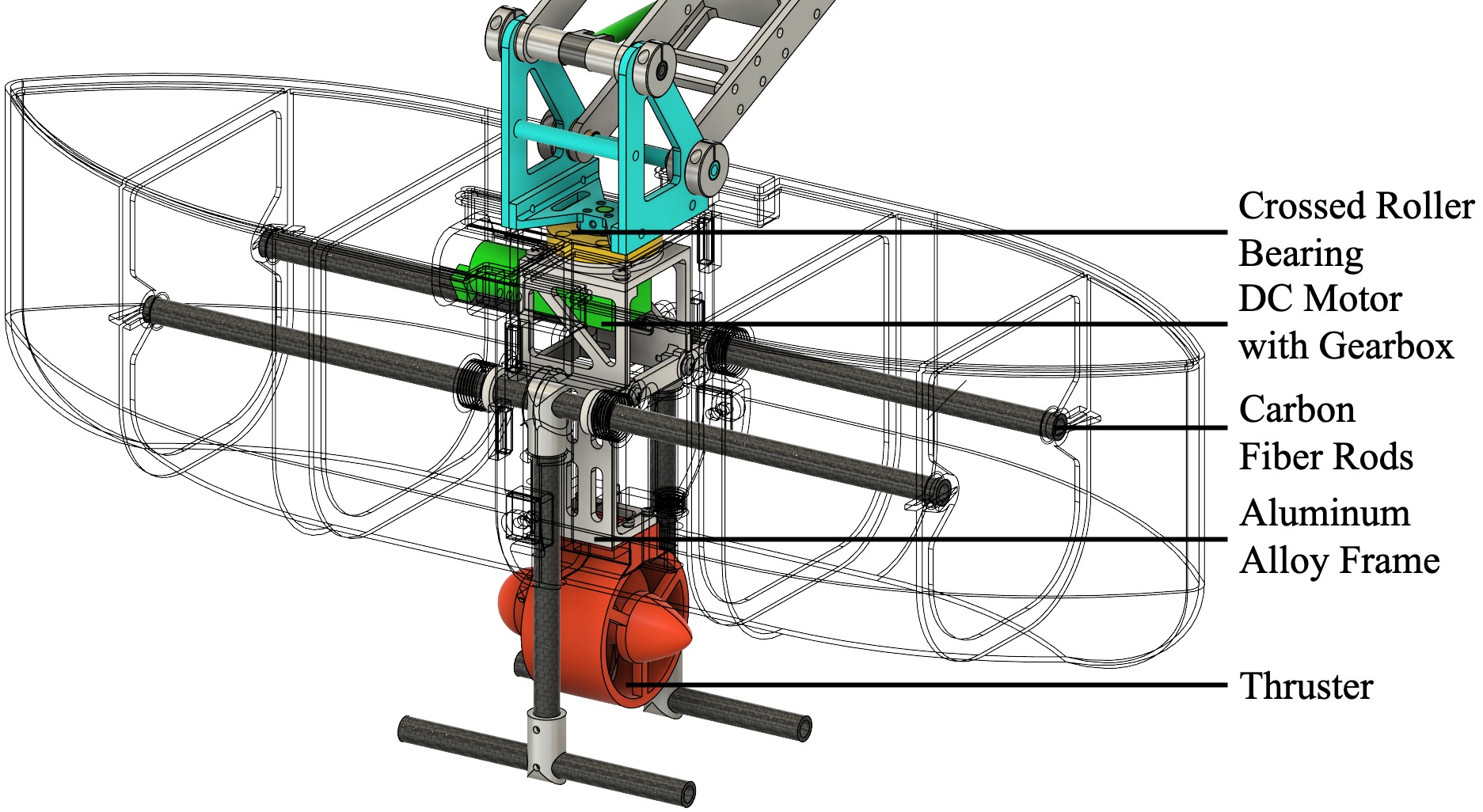}
	\caption{Main components of the outrigger hull. }
	\label{fig:outriggerhulls}
\end{figure}

\begin{figure} [htpb]
	\centering
	\subfloat[]{
		\centering
		\includegraphics[width=0.29\linewidth]{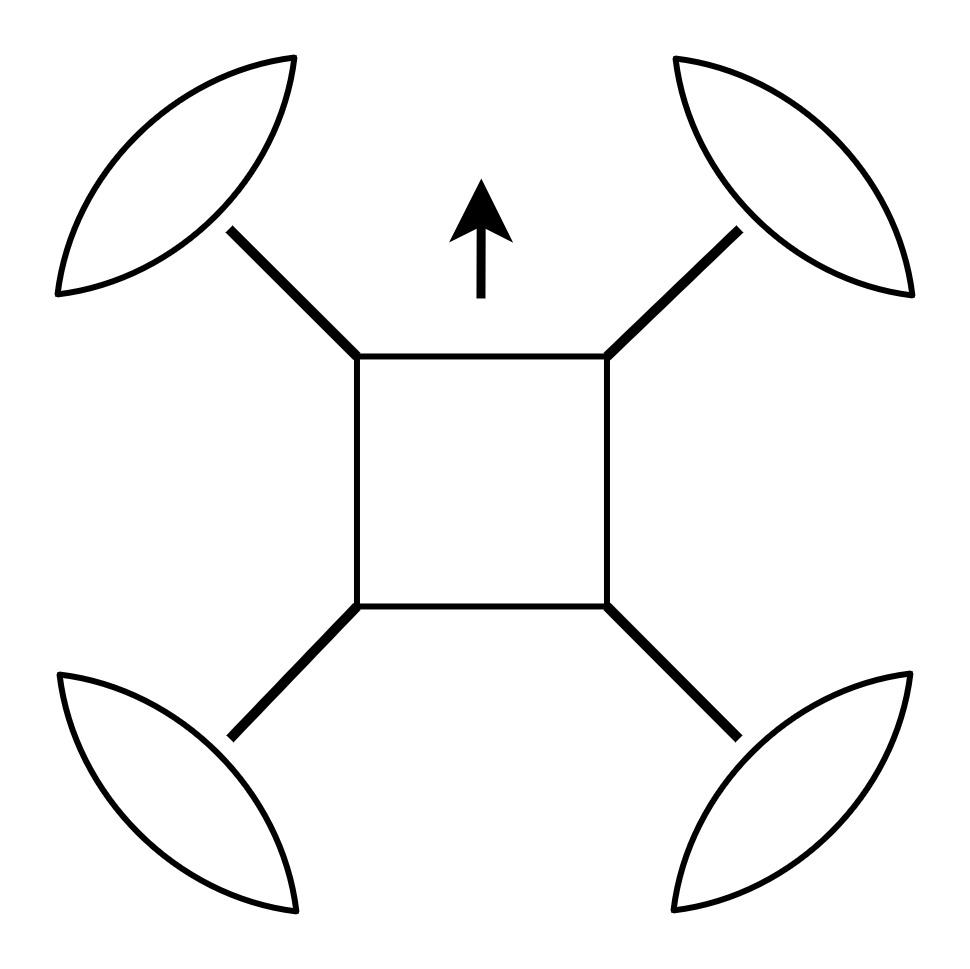}
	}
	\subfloat[]{
		\centering
		\includegraphics[width=0.22\linewidth]{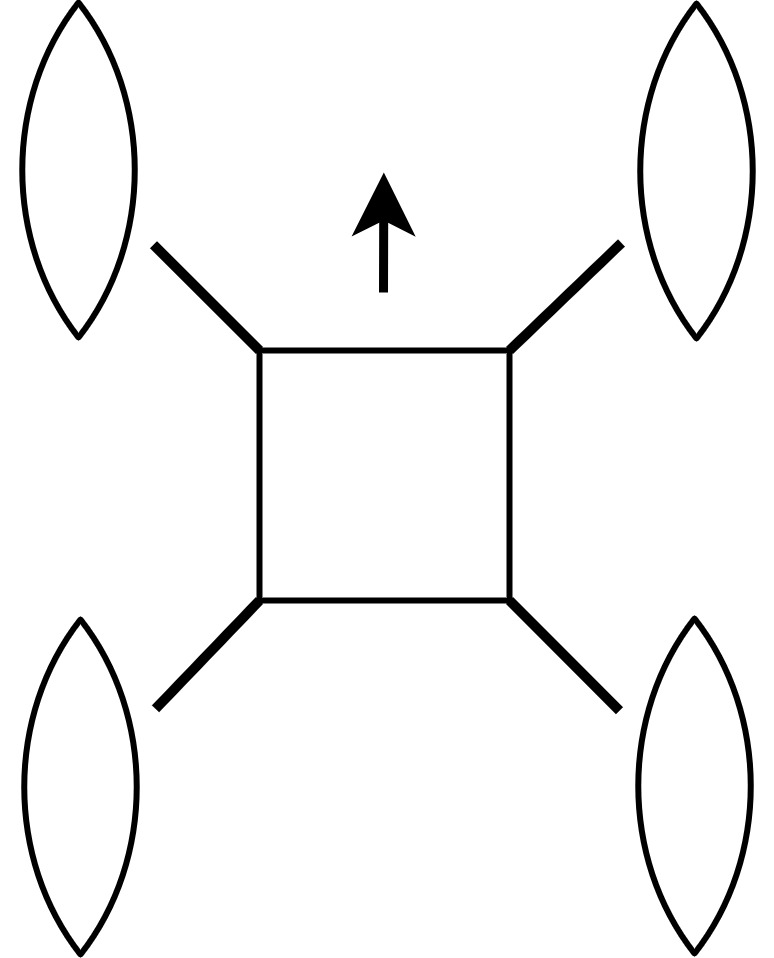}
	}
	\caption{(a) Fully-actuated mode and (b) underactuated mode. }
	\label{fig:actuation_config}
\end{figure}

Fig. \ref{fig:outriggerhulls} depicts that the ankle endo/exo-rotation (AEE) joints connect the legs and outrigger hulls through cross-roller bearings. 
Within each AEE joint, one motor with a gear reducer and encoder is utilized to orient the outrigger hull.
This motor can apply a maximum torque of $2.45 \text{N} \cdot \text{m}$ and achieve a rotary speed of 6 rpm.
The bottom thruster can deliver both forward and reverse thrust, with a maximum of about 10 N, aligning with the hull's streamline.
Thrust control is achieved by inputting pulse-width modulation (PWM) signals with different duty circles (about $6.1\% \sim 8.8\%$) to the thruster's regulator.
Coordinated with leg actions, this configuration enables QuadBoat to operate in diverse modes, including fully-actuated and underactuated, as depicted in Fig. \ref{fig:actuation_config}.
All components, including the ankle motor, thruster, carbon fiber rods (CFRs), and hull, are installed on an aluminum alloy framework, where the CFRs support the hull and protect the thruster.
To ensure balance, we assume in this paper that the outrigger hulls symmetrically position around the center of the boat, offering a total buoyancy of about 55kg.

\subsection{Electronics}

\begin{figure} [htpb]
	\centering
	\subfloat[]{
		\centering
		\includegraphics[width=0.6\linewidth]{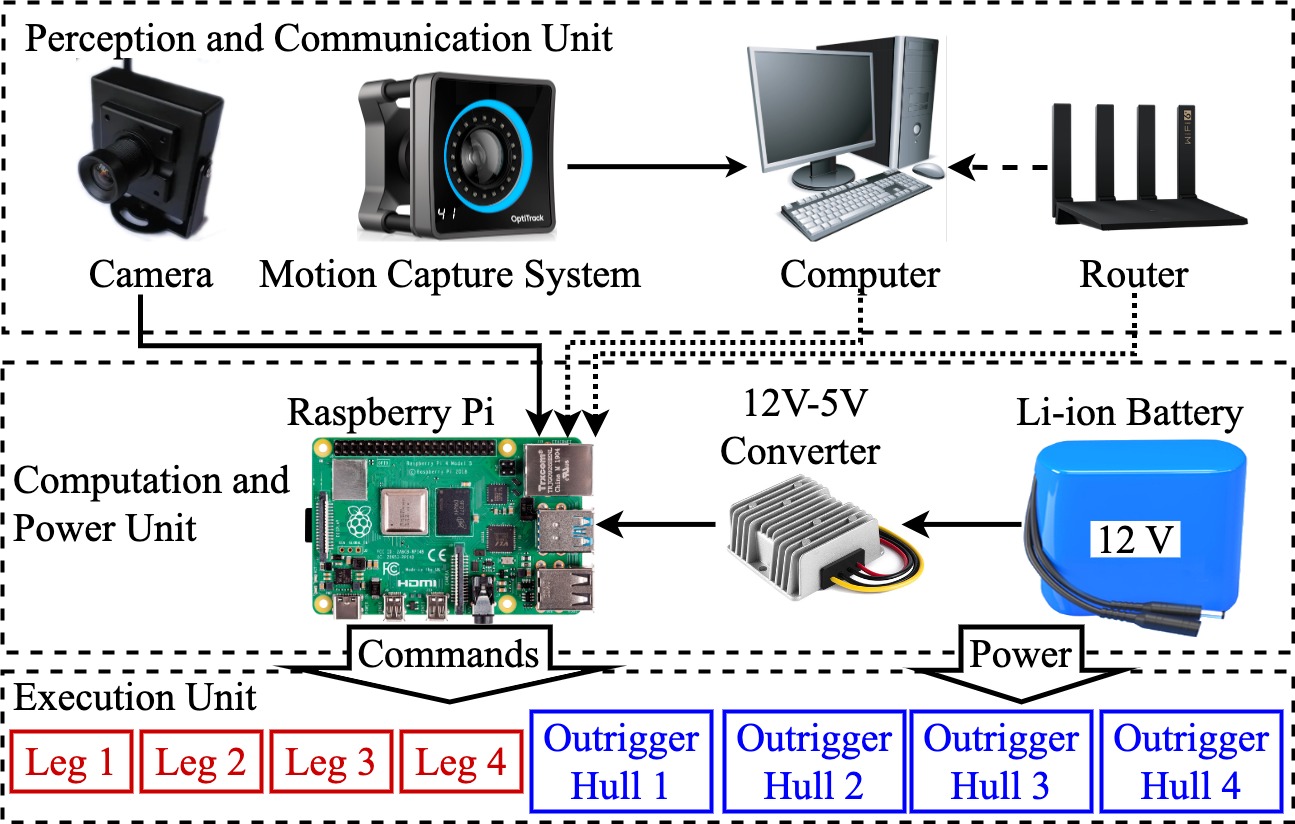}
	}
	
	\subfloat[]{
		\centering
		\includegraphics[width=0.345\linewidth]{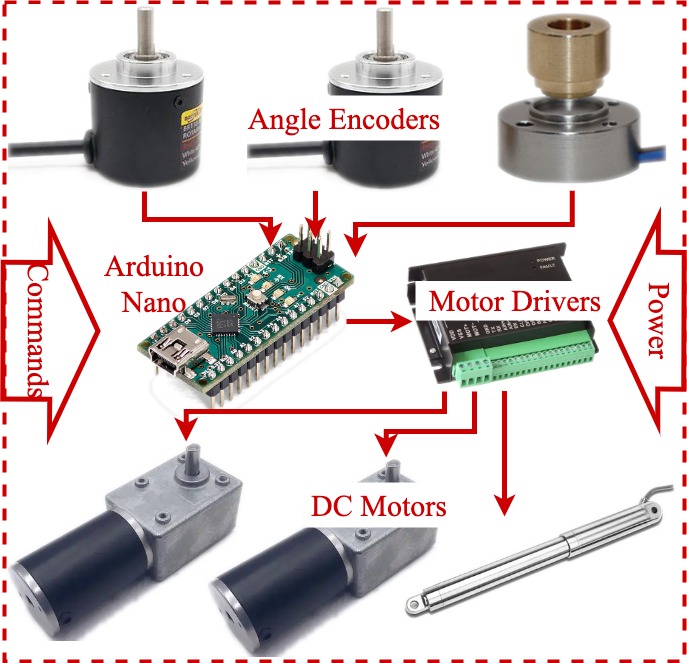}
	}
	\subfloat[]{
		\centering
		\includegraphics[width=0.3\linewidth]{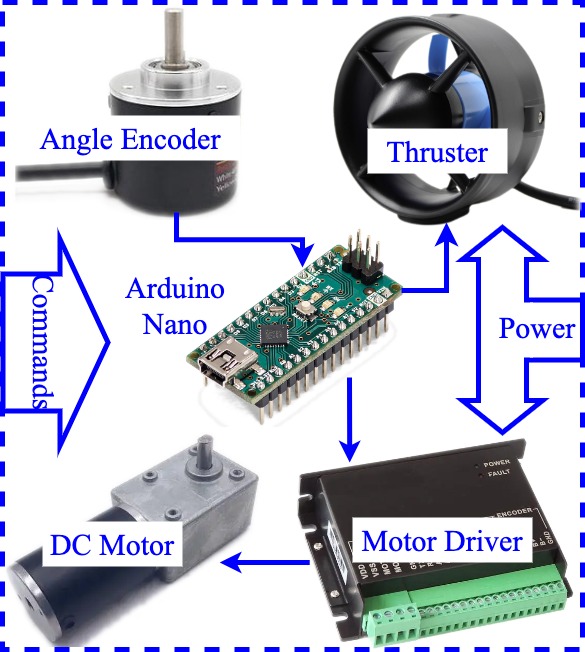}
	}
	\caption{(a) Architecture of the electronic system, (b) components of each leg, and (c) components of each outrigger hull. }
	\label{fig:electronic_system}
\end{figure}

The architecture of the electronic subsystem is shown in Fig. \ref{fig:electronic_system}. 
The main electronic components are primarily divided into the perception and communication layer, computation and power layer, and execution layer.
(1) Various sensors enable navigation and environmental perception. 
Cameras identify targets and their positions, while a motion capture system enables global localization with millimeter-level precision. 
The offshore computer communicates with QuadBoat via a wireless local area network, facilitating the transmission of positioning information and user remote control commands.
(2) The processing unit responsible for executing the navigation and control algorithms is a Raspberry Pi 4B (1.8 GHz ARM Cortex-A72, 2GB LPDDR4), from which instructions are transmitted through the serial ports to control the legs and outrigger hulls.
We employ a 12V Li-ion battery pack with a total capacity of 58800 mAh and a 12V-5V converter, serving as the power source for the motors and processing unit, respectively.
(3) Each leg utilizes three closed-loop control circuits to drive joint, comprising motors, motor drivers, and angle encoders for feedback, as depicted in Fig. \ref{fig:electronic_system} (b).
The PID controller, operated by an Arduino Nano board, ensures precise control of all joint movements.
Each outrigger hull incorporates a closed-loop motor-encoder circuit and a thruster, all controlled by an Arduino board, as illustrated in Fig. \ref{fig:electronic_system} (c).
The motor control tasks are handled by Arduino, allowing the processing unit to concentrate on communication and overall control.

\subsection{Robot Fabrication}
For the fabrication of the QuadBoat, the main hull and outrigger hulls are 3D-printed with hollow shell structures to ensure waterproofing and buoyancy, using nylon material at a density of 1130 $\rm{kg/m^3}$. 
The outrigger hulls together provide a total buoyancy of around 540 N.
Meanwhile, to enhance the structural strength and stiffness, the robot's leg and central body framework employ carbon fiber reinforced polymer (CFRP) plates for the structure and aluminum alloy for the plate connectors.
The key specifications of the QuadBoat are listed in Table \ref{tab:spec}.


\begin{table}[tbp]
	\centering
	\caption{TECHNICAL SPECIFICATIONS OF QUADBOAT}
	\label{tab:spec}
	\begin{tabular}{lll}
		\toprule[1.5pt]
		\multicolumn{1}{c}{}                                                                                      & \textbf{Items}                                                                     & \textbf{Value}          \\ \hline
		\multicolumn{1}{l|}{\multirow{3}{*}{\begin{tabular}[c]{@{}l@{}}Physical \\ characteristics\end{tabular}}} & \multicolumn{1}{l|}{Width range}                                                   & 1.188 m $\sim$ 1.85 m   \\  
		\multicolumn{1}{l|}{}                                                                                     & \multicolumn{1}{l|}{Height range}                                                  & 0.425  m $\sim$ 0.643 m \\  
		\multicolumn{1}{l|}{}                                                                                     & \multicolumn{1}{l|}{Total mass}                                                    & 25 kg                   \\ \hline
		\multicolumn{1}{l|}{\multirow{6}{*}{Performance}}                                                         & \multicolumn{1}{l|}{Max. speed}                                                    & 0.8 m/s                        \\  
		\multicolumn{1}{l|}{}                                                                                     & \multicolumn{1}{l|}{Max. steering speed} &  $\pi /3$ rad/s                      \\ 
		\multicolumn{1}{l|}{}                                                                                     & \multicolumn{1}{l|}{Capacity}                                                      & 20 kg                        \\  
		\multicolumn{1}{l|}{}                                                                                     & \multicolumn{1}{l|}{Battery life}                                                  &   2 hours (normal use)                     \\ 
		\multicolumn{1}{l|}{}                                                                                     & \multicolumn{1}{l|}{Control mode}                                                  & Autonomous/remote control                         \\
		\multicolumn{1}{l|}{}                                                                                     & \multicolumn{1}{l|}{Movement mode}                                                 & Fully-actuated/underactuated                        \\ 
		\bottomrule[1.5pt]
	\end{tabular}
\end{table}


\section{Modeling}
\label{sect:model}

\subsection{Kinematics of Legs}

\begin{figure} [htbp] 
	\centering
	\includegraphics[width=0.7\linewidth]{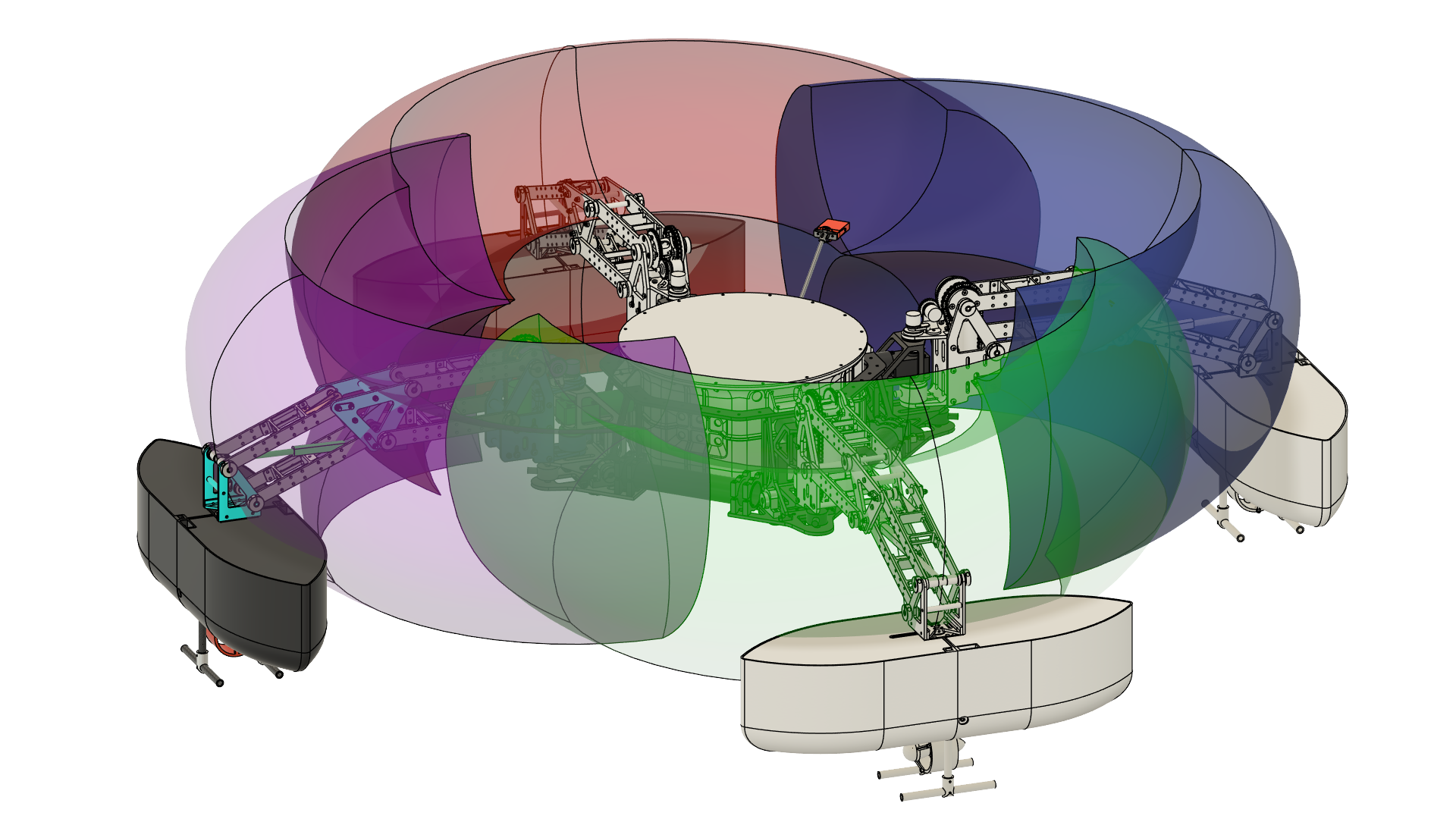}
	\caption{Overlapping workspaces of the legs. Rotation of all outrigger hulls is omitted. }
	\label{fig:LegWorkspace}
\end{figure}

\begin{figure} [htbp] 
	\centering
	\includegraphics[width=0.7\linewidth]{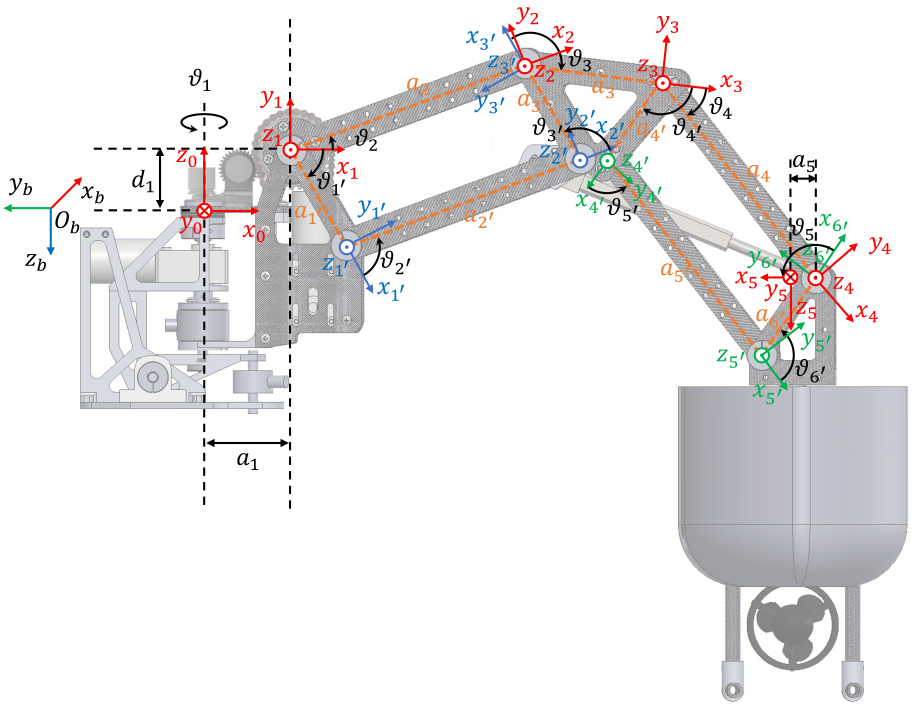}
	\caption{The $D-H$ coordinate system for one leg. }
	\label{fig:LegDHCoordinate}
\end{figure}

\begin{figure} [htpb]
	\centering
	\includegraphics[width=0.6\linewidth]{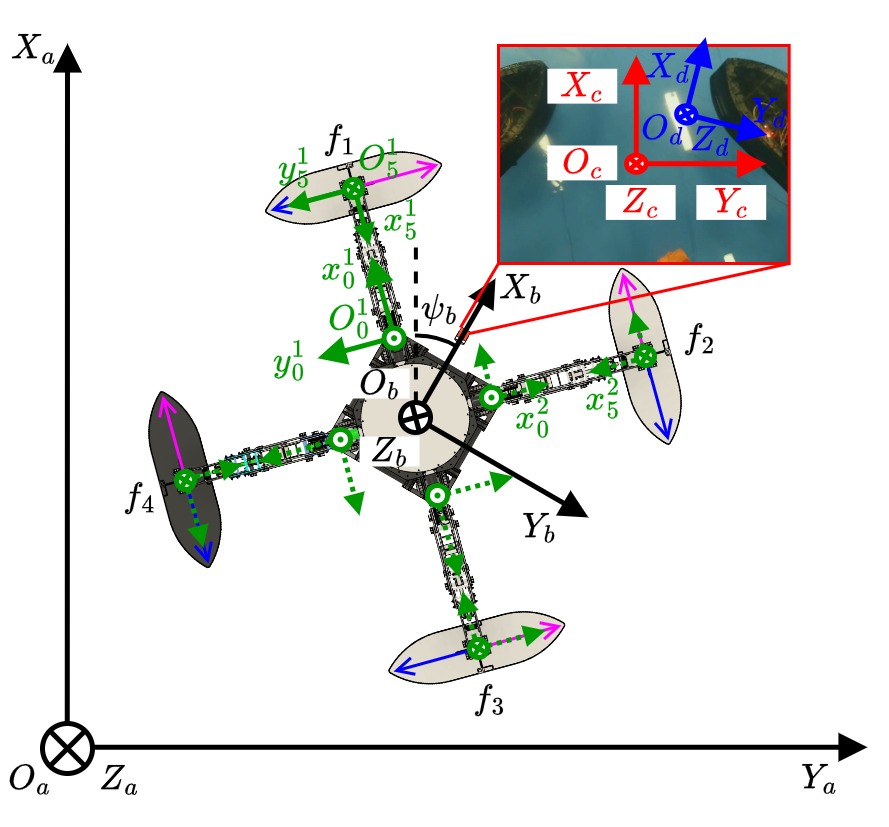}
	\caption{Coordinate system for the movement of the QuadBoat: inertial coordinate $O_a \text{-} X_a Y_a Z_a$, body coordinate $O_b \text{-} X_b Y_b Z_b$, camera coordinate $O_c \text{-} X_c Y_c Z_c$, and target coordinate $O_d \text{-} X_d Y_d Z_d$. $O_0^j \text{-} x_0^j y_0^j z_0^j$ and $O_5^j \text{-} x_5^j y_5^j z_5^j$ denote the base and end coordinates of each leg, respectively. Red arrows stand for positive propulsion forces and blue arrows stand for negative forces. }
	\label{fig:coordinate}
\end{figure}

Fig. \ref{fig:LegWorkspace} depicts the workspace of the QuadBoat's legs, where each leg produces a toroid, with overlaps between them, necessitating modeling and control of their actions to avoid interference.
Therefore, we initially establish their inverse kinematic models.
As shown in Fig. \ref{fig:LegDHCoordinate}, we establish the Denavit-Hartenberg ($D-H$) coordinate system for all four legs, since they own identical kinematic mechanisms.
Let $\{O_0^j-x_0^jy_0^jz_0^j\}\ (j=1,2,3, 4)$ denote the base frame of each leg, whose origin is located at the rotation axis of the HEE joint.
$\{O_5^j-x_5^jy_5^jz_5^j\}$ represents the frame of the leg end point (LEP).
The body coordinate $\{O_b-X_bY_bZ_b\}$ is located at the center of central body with the $X_b$ axis pointing forwad and the origin $O_b$ in the same plane as the four leg frames, as depipcted in Fig. \ref{fig:coordinate}.
The origin of four leg frames with respect to the body frame are $(b,-b,0)$, $(b,b,0)$, $(-b,b,0)$, and $(-b,-b,0)$, where $b=200\ \text{mm}$.

\begin{table}[tbp]
	\centering
	\caption{D-H PARAMETERS FOR LEGS}
	\label{tab:DHparameter}
	\begin{tabular}{cllll}
		\toprule[1.5pt]
		\textbf{Link} & $a_i $ (mm)      & $\alpha_i$ (rad)             & $d_i$ (mm)                 & $\vartheta_i$ (rad) \\
		\midrule
		1  & $a_{1} = 72 $                     & $\alpha_1 = \pi/2 $          & $d_1 = 44.71 $            &  $\vartheta_1$ \\
		2  & $a_{2} = 200 $                  & $\alpha_2=0$                  & $d_2 = 0 $                 & $\vartheta_2$ \\
		3  & $a_{3} = 110.72  $            & $\alpha_3 = 0 $               & $d_3 = 0 $                 & $\vartheta_3 \approx 2.225$ \\
		4  & $a_{4} = 200 $                 & $\alpha_4= 0$                 & $d_4 = 0 $                 & $\vartheta_4$ \\
		5  & $a_{5} = 200$                  & $\alpha_5= -\pi/2$          & $d_5 = 0 $                  & $\vartheta_5 = 25\pi/36$ \\
		1'  & $a_{1^\prime}= 88 $       & $\alpha_{1^\prime}= 0$  & $d_{1^\prime} = 0 $  & $\vartheta_{1^\prime}=-\pi/3$ \\
		2'  & $a_{2^\prime} = 200 $   & $\alpha_{2^\prime}= 0$ & $d_{2^\prime} = 0 $  & $\vartheta_{2^\prime}=\vartheta_2-\vartheta_{1^\prime}$ \\
		3' & $a_{3^\prime} =88 $       & $\alpha_{3^\prime}= 0$ & $d_{3^\prime} = 0 $  & $\vartheta_{3^\prime}=\pi-\vartheta_{2^\prime}$ \\
		4' & $a_{4^\prime}=75.46 $   & $\alpha_{4^\prime}= 0$ & $d_{4^\prime} = 0 $ & $\vartheta_{4^\prime} \approx 2.052$ \\
		5' & $a_{5^\prime}=200 $      & $\alpha_{5^\prime}= 0$ & $d_{5^\prime} = 0 $  & $\vartheta_{5^\prime}=\vartheta_4-\vartheta_{4^\prime}$ \\
		6' & $a_{6^\prime}=75.46 $  & $\alpha_{6^\prime}= 0$ & $d_{6^\prime} = 0 $  & $\vartheta_{6^\prime}=\pi-\vartheta_{5^\prime}$ \\
		\bottomrule[1.5pt]
	\end{tabular}
\end{table}

In this section, the superscript $j$ of each parameter is ommitted to facilletate analysis.
The $D-H$ parameters ($a_i$,  $\alpha_i$, $d_i$, $\vartheta_i$) of the proposed $D-H$ system are listed in Table \ref{tab:DHparameter}.
Among these parameters, $\vartheta_1$, $\vartheta_2$, and  $\vartheta_4$ are the rotational angle driven by actuators of HEE, HFE, and KFE joints, respectively.
The leg is formed by sequentially connecting three links (1,3,5) and two closed chains with parallelogram structures from the LEP frame to the leg base frame.
We can derive the homogeneous transformation matrices for the three open chains (frame $1$ to $0$, $3$ to $3^\prime$, and $5$ to $6^\prime$) as follows.


\begin{equation} \label{eq:TransMatrix01}
	{\mathbf{A}_{1}^{0}}(\vartheta_1) = 
	\left [{\begin{matrix}
			{{c}_{1}}&0&{{s}_{1}}&{{a}_{1}}{{c}_{1}}\\
			{{s}_{1}}&0&-{{c}_{1}}&{{a}_{1}}{{s}_{1}}\\
			0&1&0&{{d}_{1}}\\
			0&0&0&1
	\end{matrix}}\right ],
\end{equation}

\begin{equation}
	{\mathbf{A}_{{{3}}}^{3^{'}}}(\vartheta_3) =
	\left [{\begin{matrix}
			{{c}_{3}}&{{s}_{3}}&0&{{a}_{3}}{{c}_{3}}\\
			{{s}_{3}}&-{{c}_{3}}&0&{{a}_{3}}{{s}_{3}}\\
			0&0&1&0\\
			0&0&0&1
	\end{matrix}}\right ],
\end{equation}
and 
\begin{equation}
	{\mathbf{A}_{5}^{6^\prime}}(\vartheta_5)=
	\left [{\begin{matrix}
			{{c}_{5}}&0&{-{s}_{5}}&{{a}_{5}}{{c}_{5}}\\
			{{s}_{5}}&0&{{c}_{5}}&{{a}_{5}}{{s}_{5}}\\
			0&-1&0&0\\
			0&0&0&1
	\end{matrix}}\right ],
\end{equation}
where $s_i= \sin \vartheta_i$, and $c_i= \cos \vartheta_i$, which remains hereafter.

According to \cite{Siciliano2009}, the transformation matrices (from $3^\prime$ to $1$ and $6^\prime$ to $3$) of two closed chains formed by the parallelogram structures can be derived as follows.

\begin{equation}
	{\mathbf{A}_{{{3}^{'}}}^{1}}(\vartheta_2)=
	\left [{\begin{matrix}
			{-{c}_{{1}^{'}}}&{{s}_{{1}^{'}}}&0&{\xi_{c_2}}\\
			{-{s}_{{1}^{'}}}&{-{c}_{{1}^{'}}}&0&{\xi_{s_2}}\\
			0&0&1&0\\
			0&0&0&1\end{matrix}}\right ]
\end{equation}
and 
\begin{equation} \label{eq:TransMatrix36}
	{\mathbf{A}_{{{6}^{'}}}^{3}}(\vartheta_4)=
	\left [{\begin{matrix}{-{c}_{{4}^{'}}}&{{s}_{{4}^{'}}}&0&\xi_{{c}_{4}}\\
			{-{s}_{{4}^{'}}}&{-{c}_{{4}^{'}}}&0&\xi_{{s}_{4}}\\
			0&0&1&0\\
			0&0&0&1\end{matrix}}\right ]
\end{equation}
where ${\xi_{c_2}}=({a}_{{1}^{'}}-{a}_{{3}^{'}}){{c}_{1^\prime}}+{a}_{{2}^{'}}{c}_{2}$,$\xi_{{s}_{4}}=({a}_{{1}^{'}}-{a}_{{3}^{'}}){{s}_{1^\prime}}+{a}_{{2}^{'}}{s}_{2}$, $\xi_{c_4}={({a}_{{4}^{'}}-{{a}_{{6}^{'}}})}{{c}_{{4}^{'}}}+{{a}_{{{5}^{'}}}}{{c}_{4}}$, and $\xi_{s_4}={({a}_{{4}^{'}}-{{a}_{{6}^{'}}})}{{s}_{{4}^{'}}}+{{a}_{{{5}^{'}}}}{{s}_{4}}$.


Based on the above Eq. (\ref{eq:TransMatrix01}) -  (\ref{eq:TransMatrix36}), the transformation matrix from the end frame to the base frame is expressed as
	\begin{equation} \label{eq:TransMatrix05}
		\begin{aligned}
			{\mathbf{A}_{5}^{0}}
			&={\mathbf{A}_{1}^{0}}{\mathbf{A}_{3^\prime}^{1}}{\mathbf{A}_{3}^{3^\prime}}{\mathbf{A}_{{{6}^{'}}}^{3}}{\mathbf{A}_{5}^{6^\prime}} 
			=\left [{\begin{matrix}
					\mathbf{R}_5^0 & \bm{P}_5^0 \\
					0 & 1
			\end{matrix}}\right ] \\
			&=\left [{\begin{matrix}
					{{c}_{1}{c}_{{1}^{\prime}3\bar{4^\prime}\bar{5}}}&-{s}_{1}&{{c}_{1}{s}_{{1}^{\prime}3\bar{4^\prime}\bar{5}}}&{{c}_{1}(a_5{c}_{{1}^{\prime}3\bar{4^\prime}\bar{5}}+\xi_{c_{24}s_4})}\\
					{{s}_{1}{c}_{{1}^{\prime}3\bar{4^\prime}\bar{5}}}&{c}_{1}&{{s}_{1}{s}_{{1}^{\prime}3\bar{4^\prime}\bar{5}}}&{{s}_{1}(a_5{c}_{{1}^{\prime}3\bar{4^\prime}\bar{5}}+\xi_{c_{24}s_4})}\\
					{s}_{{1}^{\prime}34^\prime5}&0&{c}_{{1}^{\prime}34^\prime5}&a_5{s}_{{1}^{\prime}3{4^\prime}{5}}+\xi_{s_{24}c_4}\\
					0&0&0&1
			\end{matrix}}\right ] \\
		\end{aligned},
	\end{equation}
where ${c}_{{1}^{\prime}3\bar{4^\prime}\bar{5}}=\cos(\xi_{1^\prime}+\xi_3-\xi_{4^\prime}-\xi_5)$ and ${s}_{{1}^{\prime}3\bar{4^\prime}\bar{5}}=\sin(\xi_{1^\prime}+\xi_3-\xi_{4^\prime}-\xi_5)$ with the overline of the subscript indicating subtraction; $\xi_{c_{24}s_4}=\xi_{c2}-{c}_{1^\prime3}\xi_{c4}-s_{1^\prime3}\xi_{s4}-a_3c_{1^\prime3}+{a}_{1}$ and $\xi_{s_{24}c_4}=\xi_{s2}-{c}_{1^\prime3}\xi_{s4}-s_{1^\prime3}\xi_{c4}-{{a}_{3}}{{s}_{1^\prime3}}+d_1$.
$\bm{P}_5^0$ is the origin position of the end frame which the LEP belongs to.
Once $\bm{P}_5^0$ is given, the rotational angles $\vartheta_1,\vartheta_2,\vartheta_4$ can be solved.

\subsection{Propulsion Model}
The posture of each leg, especially the angles of HEE and AEE joints, determines the thrust direction of the outrigger hull mounted at the end, thereby influencing the robot's navigation.
We can represent the base frame $\{O_0^j-x_0^jy_0^jz_0^j\}$ of the HEE joint and the end frame $\{O_5^j-x_5^jy_5^jz_5^j\}$ of the AEE joints for the $j$-th leg in the overall coordinate system of the QuadBoat, as depicted in Fig. \ref{fig:coordinate}.
Let $\vartheta_0^j$ and $\vartheta_a^j$ denote the rotation angles with respect to $\{O_0^j-x_0^jy_0^jz_0^j\}$ and $\{O_5^j-x_5^jy_5^jz_5^j\}$, respectively. 
In each leg $j$, these two angles can be measured by the encoders installed at the HEE and AEE joints, both set to 0 in the fully-actuated configuration illustrated in Fig. \ref{fig:coordinate}.
Therefore, we can derive the angle $\theta_j$ between the thrust direction and the front of QuadBoat, i.e., the x-axis of the body frame, as follows,
\begin{equation} \label{eq:thrust_angle}
	\theta_j = {\vartheta}^j_a-{\vartheta}^j_1+\zeta_j ,
\end{equation}
where the constants $\zeta_j$ are $\zeta_1 = \zeta_3 =  \pi/4, \zeta_2 = \zeta_4 = - \pi/4$.
Using $\bm{u}=\left[f_1, f_2, f_3, f_4 \right]^T$ to denote the propulsion vector with $f_j$ being the propulsion generated by each thruster, the applied force and moments  $\bm{F}$ can be computed by
\begin{equation}
	\bm{F} = \mathbf{E} \bm{u}=
	\left[
	\begin{array}  {cccc}
		\cos{\theta_1} & \cos{\theta_2} & \cos{\theta_3} & \cos{\theta_4} \\ 
		\sin{\theta_1}  & \sin{\theta_2}  & \sin{\theta_3}   & \sin{\theta_4} \\ 
		L_1 & -L_2 & -L_3 & L_4
	\end{array}
	\right]
	\left[
	\begin{array}  {cccc}
		f_1 \\ f_2 \\ f_3 \\ f_4
	\end{array}
	\right],
\end{equation}
where $\mathbf{E}$ is mapping matrix and $L_j = (l_c + l_b^j \cos{\vartheta^j_1}) \cos{\vartheta^j_a}, \, ( j=1,2,3,4)$ is the moment arm of the propulsion $f_j$ with respect to the USV body center $O_b$, with $l_c$ and $l_d$ denoting the distance from $O_0^j$ to $O_b$ and $O_5^j$ in the $X_bO_bY_b$ plane, respectively.
To ensure balanced movement, we assume QuadBoat operates only in two symmetrical modes: underactuated and fully-actuated, as shown in Fig. \ref{fig:actuation_config}.
For control simplicity in the first step, we only consider the fully-actuated configuration, where $\vartheta_1^j=\vartheta_a^j=0$ and  $L_j=637\ \text{mm}$ with $l_c = 275\ \text{mm}$ and $l_b = 362 \ \text{mm}$.
The matrix $E$ becomes
\begin{equation}
	\mathbf{E} =
	\left[
	\begin{array}  {cccc}
		\sqrt{2}/2  & \sqrt{2}/2    &\sqrt{2}/2    & \sqrt{2}/2 \\ 
		\sqrt{2}/2  & -\sqrt{2}/2  & \sqrt{2}/2   & -\sqrt{2}/2 \\ 
		L_1 & -L_2 & -L_3 & L_4
	\end{array}
	\right].
\end{equation}
Once $\bm{F}$ is computed, the propulsion froces of all thrusters can be obtained by
\begin{equation}
	\bm{u} =  \mathbf{E}^T (\mathbf{E} \mathbf{E}^T )^{-1}  \bm{F}.
\end{equation}
%

\section{Control Strategy}
\label{sect:control}

\subsection{Control Frame}
\begin{figure} [htbp] 
	\centering
	\includegraphics[width=0.7\linewidth]{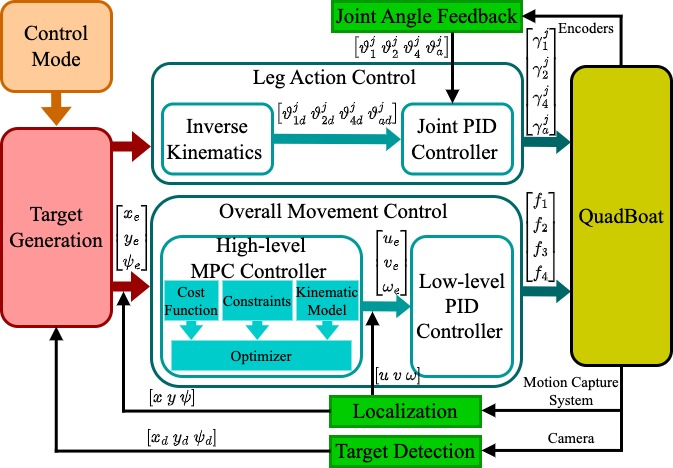}
	\caption{The control frame for QuadBoat. }
	\label{fig:controlframe}
\end{figure}

Fig. \ref{fig:controlframe} demonstrates the overall framework of the QuadBoat's control software implemented in the robot operation system (ROS).
The control mode offers two options: remote control and autonomous control, which are determined by the user.
The target generation module assigns target points to both the leg action and overal movement based on the user-selected task, and generates a series of interim target points through linear interpolation from the current position to the target position.
Then, the leg action and overall movement are independently controlled to track these interim targets.
Specifically, we present a PID-based inverse kinematic control strategy for the leg action and an MPC controller for the overall movement propelled by the propulsion system, both of which are explained in the followings.

\subsection{Joint Control of Legs}
We represent the target LEPs for each leg $j$ in the body frame $\{O_b-X_bY_bZ_b\}$ as $\bm{P}_d^j \in \mathbb{R}^3$.
Then, given the targets $\bm{P}_d$ and orientation $\theta_j$ of AEE joint, the desired rotational angles of HEE, HFE, KFE, and AEE joints, $\bm{\vartheta}_d^j = \left[ \vartheta_{1d}^j, \vartheta_{2d}^j, \vartheta_{4d}^j, \vartheta_{ad}^j \right]^T$, can be solved from Eq. (\ref{eq:TransMatrix05}) and (\ref{eq:thrust_angle}) which is
\begin{equation}
	\bm{P}_5^{oj} =  \bm{P}_d^j - \left[0,0,z_d^j\right], \quad
	{\vartheta}^j_a = \theta_j + {\vartheta}^j_1 - \zeta_j ,
\end{equation}
where $z_d^j$ is the z-coordinate of LEP in the leg end frame $\{O_5^j-x_5^jy_5^jz_5^j\}$.
It's worth noting that the action of each joint is constrained within the following ranges,
\begin{equation}
	\vartheta_1^j \in [-\frac{\pi}{2},\frac{\pi}{2}],  \,
	\vartheta_2^j \in [-\frac{\pi}{6}, \frac{5\pi}{12}],  \, 
	\vartheta_4^j \in [-\frac{\pi}{3}, 0], \,
	\vartheta_a^j \in [-\frac{\pi}{2},\frac{\pi}{2}].
\end{equation}
Let $\bm{\vartheta}^j  \left ( t \right ) = \left[ \vartheta_{1}^j, \vartheta_{2}^j, \vartheta_{4}^j, \vartheta_{a}^j \right]^T$ denote the joint angle feedback measured by encoders  at time $t$ and $\bm{\vartheta}_e \left ( t \right )  = \bm{\vartheta}^j  \left ( t \right ) - \bm{\vartheta}_d^j  \left ( t \right )$ be the errors.
The control commands to be input to the actuator regulators, $\bm{C} \left ( t \right ) = \left[ \gamma^j_{1}, \gamma^j_{2}, \gamma^j_{4}, \gamma^j_{a} \right]^T$ , can be computed by a discrete PID controller, that is
\begin{equation}
	\bm{C} \left ( t \right ) = \mathbf{K}_P  \bm{\vartheta}_e \left ( t \right ) + \mathbf{K}_I \sum_{\tau=0}^t  \bm{\vartheta}_e\left ( \tau \right ) + \mathbf{K}_D \left ( \bm{\vartheta}_e\left ( t \right ) - \bm{\vartheta}_e\left ( t - 1 \right ) \right ) ,
\end{equation}
where $ \mathbf{K}_P,  \mathbf{K}_I,  \mathbf{K}_D$ are the diagnol gain matrices, whose values are identical for all legs. 
Notably, all PID controllers in this paper are tuned through the Ziegler?Nichols method \cite{McCormack1998}, of which the gains are exhibited in Table \ref{tab:pid_gains}.


\begin{table}[tbp]
	\centering
	\caption{PID CONTROLLER GAINS}
	\label{tab:pid_gains}
		\begin{tabular}{cccccccc}
			\toprule[1.5pt]
			& \multicolumn{4}{c|}{\textbf{Leg joint control}}    & \multicolumn{3}{c}{\textbf{Movement control}}       \\ \hline
			& HEE & HFE & KFE & \multicolumn{1}{c|}{AEE} & $X_b$ & $Y_b$  & $\psi_b$  \\ \hline
			$K_P$           & 20   & 50     & 250        & 15            & 7015             & 8520        & 2095    \\
			$K_I$            & 1.1     & 0.3        & 0.6            & 0.8           & 12                  & 13             & 8    \\
			$K_D$          & 0.7      & 0.2       & 0.4            & 0.5            & 8                   & 8               & 5      \\  
			\bottomrule[1.5pt]
		\end{tabular}
\end{table}

\subsection{Movement Control}
We have chosen MPC as the high-level controller for the QuadBoat due to its advantages of high precision and efficiency on the omnidirectional USVs \cite{Zhang2022a}. 
Fig. \ref{fig:coordinate} illustrates the QuadBoat in a two-dimensional inertial frame $\{O_a-X_aY_aZ_a\}$, where the body frame $\{O_b-X_bY_bZ_b\}$ is set at the body center with $X_b$ toward the front, and $\{O_c-X_cY_cZ_c\}$ and $\{O_d-X_dY_dZ_d\}$ represent the camera frame and the target frame, respectively.
Since only the symmetric action is involved, the origin $O_b$ of the body frame is considered as the center of mass (COM).
The position and orientation of the QuadBoat in $\{O_a-X_aY_aZ_a\}$ relative to the COM are defined as $\bm{\eta}=\left[x,y,\psi \right]^T $, while the robot's velocity in $\{O_b-X_bY_bZ_b\}$ is defined as $\bm{v}=\left[u,v,\omega \right]^T $.
Therefore, the kinematics of the QuadBoat movement can be expressed as
\begin{equation}
	\dot{\bm{\eta}}= \mathbf{R}(\bm{\eta})\bm{v},
\end{equation}
where $\mathbf{R} \left ( \bm{\eta} \right )$ is the transformation matrix converting a state vector from body frame to inertial frame, which is
\begin{equation}
	\mathbf{R}(\bm{\eta})= {\left[ {\begin{array}{*{20}{c}}
				{\cos{\psi}}&
				{-\sin{\psi}}&
				0\\
				
				{\sin{\psi}}&
				{\cos{\psi}}&
				0\\
				
				0&
				0&
				1
		\end{array}} \right] }.
\end{equation}

Let $\bm{\eta}_d$ and  $\bm{v}_d$ describe the desired trajectory.
We define the position error in body frame as
\begin{equation}
	\bm{\eta}_e = \mathbf{R}^T(\bm{\eta})(\bm{\eta}-\bm{\eta}_d)
\end{equation}
and its derivative
\begin{equation}
	\dot{\bm{\eta}_e} 
	=\mathbf{f}_e(\bm{\eta}_e,\bm{v})
	= \bm{v}-\mathbf{R}^T(\bm{\eta}_e)\bm{v}_d
	-\mathbf{S}\left( \left[ 0,	0,	\omega \right] ^T  \right) \bm{\eta}_e ,
\end{equation}
where $\mathbf{S}\left( \bm{a} \right) $ stands for the skew-symmetric matrix that verifies $\mathbf{S}\left( \bm{a} \right) \bm{b} = \bm{a} \times \bm{b} $ .
Then, the discrete-time difference equation of robot motion is derived using the forward Euler discretization, yielding
	\begin{equation}
		\bm{\eta}_{ek+1}
		\approx
		\bm{\eta}_{ek}+ T_s \mathbf{f}_e(\bm{\eta}_{ek}, \bm{v}_k )
		=
		\mathbf{f}_k \left(\bm{\eta}_{ek}, \bm{v}_k \right),
	\end{equation}
	with $T_s$ being the sampling time.
	
	Next, we can formulate the MPC problem to obtain the open-loop optimal control sequence within every horizon $T$.
	Let $N$ be the prediction steps in each horizon, $\bm{V} = \left[ \bm{v}_0, \ldots, \bm{v}_{N-1} \right]^T$ the control sequence, and $\bm{H} = \left[\bm{\eta}_{e0}, \ldots, \bm{\eta}_{eN} \right]^T$ the state sequence vector after inputing that control sequence.
	The model constraint within each horizon of $N$ steps can be written as
	\begin{equation}
		\mathbf{f}_M (\bm{H}, \bm{V}) =
		\left[\begin{array}{c}
			\mathbf{f}_0-\bm{\eta}_{e1} \\
			\vdots \\
			\mathbf{f}_{N-1}-\bm{\eta}_{eN} 
		\end{array}\right]
		=\mathbf{0}.
	\end{equation}
	
	Then, the optimization problem of the MPC controller can be formulated as
	\begin{equation} \label{eq:optimization}
		\begin{aligned}
			\underset{ \bm{V} }{\min}  \quad
			&\mathbf{J}\left( \bm{H}, \bm{V} \right) 
			\\
			\text{s}.\text{t}. \quad
			& \bm{H}  \in \overline{ \mathfrak{ H } },\ 
			\bm{V} \in \overline{ \mathfrak{V} }, 
			\\
			& \mathbf{f}_M (\bm{H}, \bm{V}) = \mathbf{0} ,
		\end{aligned}
	\end{equation}
	where $ \overline{ \mathfrak{ H } } = \left\{ \bm{H}: \bm{\eta}_{ei} \in \mathcal{H}, \, \forall {i=0, \ldots, N}\right\}$ and 
	$ \overline{ \mathfrak{ V } } = \left\{ \bm{U}: \mathbf{v}_i \in \mathcal{V}, \, \forall {i=0, \ldots, N-1}\right\}$ are the saturation constraints for the state and control sequences with $\mathcal{H} = \left[ \bm{\eta}_{min}, \bm{\eta}_{max} \right] $ and $\mathcal{V} = \left[ \bm{v}_{min}, \bm{v}_{max} \right] $ 
	being the feasible sets of states and control inputs of the system, respectively.
	$ \mathbf{J}\left( \bm{H}, \bm{V} \right) $ denotes the cost function, defined as 
	\begin{equation}
		\mathbf{J}\left( \bm{H}, \bm{V} \right) 
		= \bm{\eta}_{eN}^T \mathbf{Q}_N  \bm{\eta}_{eN}
		+ \sum_{i = 0}^{N-1}{
			\bm{\eta}_{ei}^T \mathbf{Q}_i  \bm{\eta}_{ei}  +
			\bm{v}_{i}^T \mathbf{Q}_v  \bm{v}_{i} 
		},
	\end{equation}
	where $\mathbf{Q}_i$ and $\mathbf{Q}_v$ are positive definite matrices used to penalize the state and control deviations, and $\mathbf{Q}_N$ is the terminal penalty matrix enhancing the NMPC algorithm.
	These weighting matrices are defined as follows:
	\begin{equation}
		\begin{aligned}
			\mathbf{Q}_i &= \mathrm{diag}\left( 4000,4000,1000 \right) , \\
			\mathbf{Q}_v &= \mathrm{diag}\left( 0.01,0.01,0.01 \right),
		\end{aligned}
	\end{equation}
	and $\mathbf{Q}_N$ is set to the same values as $\mathbf{Q}$. 
	The CasADi solver \cite{andersson2019casadi}, a nonlinear optimization tool, can efficiently solve this problem within 0.1 seconds in each control cycle, providing an optimal control sequence $\bm{V}_d^\prime = \left[ \bm{v}_{d0}, \ldots, \bm{v}_{dN-1} \right]^T$, with the first solution $\bm{v}_{d0} = \left[u_d, v_d, \omega_d \right]^T$ being executed.
	As the low-level control, we utilize three PID controllers in different coordinate directions to track these desired velocities, with their gains listed in Table \ref{tab:pid_gains}.
	
	
	The state feedback for the motion controllers can be obtained through two methods: one uses a motion capture system (or GPS) to provide feedback on the current position, with user-defined target points; the other uses a camera to provide feedback on the error between the current position and the target. 
	For the latter approach, the motion controllers aim to overlap the target point with the origin in the camera frame.
	We begin by transforming the target points from pixel coordinates to real-world coordinates.
	Let $\bm{P}_r = \left[ x_r,y_r,z_r,1 \right]^T$ and $\bm{P}_c = \left[ x_c,y_c,1 \right]^T$ represent the target point in the real-world and pixel coordinates, respectively. 
	Their relationship can be expressed as follows:
	\begin{equation}
		\bm{P}_r = \frac{1}{s_r} \mathbf{K}_c^{-1} \bm{P}_c ,
	\end{equation}
	where $s_r$  is the scale factor and $\mathbf{K}_c$ denotes the  camera intrinsic matrix.
	Note that the target orientation in the real world, denoted as $\psi_r$, is identical to that in pixel coordinates, denoted as $\psi_c$.
	Based on this principle, we use AprilTag to detect the target's position $\left[x_r,y_r\right]$ and orientation $\psi_r$, which are also the errors between the target position and the origin of the camera frame.
	Finally, we can convert these errors into the tracking errors in the QuadBoat's body frame, as follows:
	\begin{equation}
		\begin{footnotesize}
			\left[
			\begin{array}  {c}
				x_e \\ y_e \\ \psi_e \\
			\end{array}
			\right]
			=
			\left[
			\begin{array}  {ccc}
				-\cos{\phi_a}    &\sin{\phi_a}     & 0 \\ 
				\sin{\phi_a}  & \cos{\phi_a}   & 0 \\ 
				0 & 0 & -1
			\end{array}
			\right]
			\left[
			\begin{array}  {c}
				\delta \\ 0 \\ \psi_r \\
			\end{array}
			\right] -
			\left[
			\begin{array}  {c}
				l_{cb} \\ 0 \\ 0 \\
			\end{array}
			\right],
		\end{footnotesize}
	\end{equation}
	where $\phi_a=\psi_r - \arctan{\frac{y_r}{x_r + l_{cb}}}$ represents the difference between the target orientation and the line extending from the body center to the target point, and
	$\delta = \lVert \left[x_r,y_r\right]^T -   \left[l_{cb},0\right]^T \rVert_2$ indicates the distance from the target position to the body center,  with $l_{cb}= 0.42 \text{m}$ being the distance from the camera to the body center.
	The velocity error feedback can be obtained by differentiating the position error 
	$\left[ \dot x_e \\ \dot y_e \\ \dot \psi_e \\ \right]$.

	\begin{figure} [htbp] 
		\centering
		\includegraphics[width=0.6\linewidth]{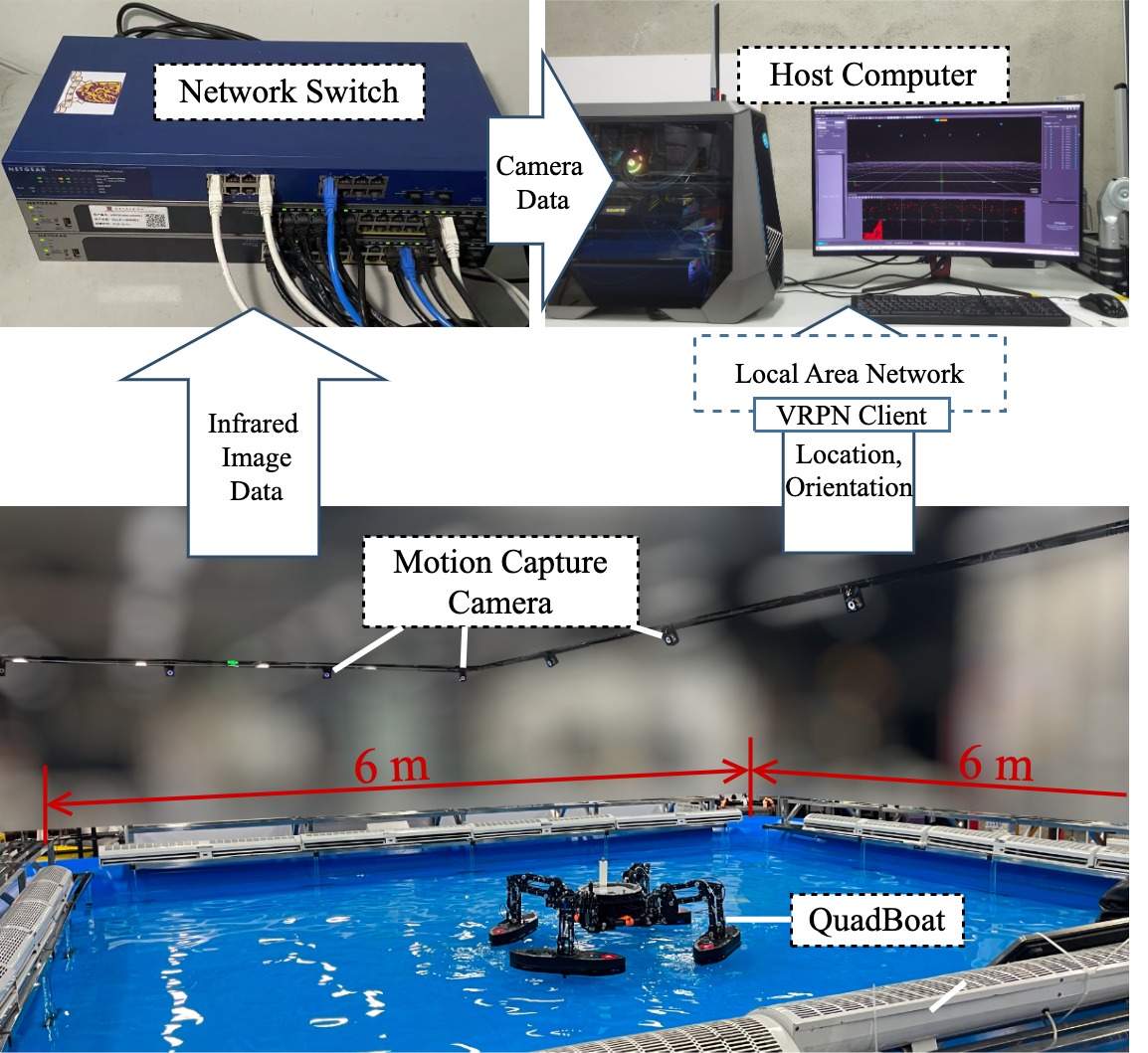}
		\caption{Experiment site setup. }
		\label{fig:exp_setup}
	\end{figure}

\section{Experiments}
\label{sect:experiment}
This section encompasses a series of experiments successively conducted in a $6 \text{m} \times 6 \text{m}$ pool, including maneuverability demonstration, leg action tracking, trajectory tracking, visual feedback tracking, and object pickup, to demonstrate the QuadBoat's tracking performance and practical utility.
The experimental setup is exhibited in Fig. \ref{fig:exp_setup}.
Firstly, the maneuverability demonstration experiment has shown that QuadBoat can adapt to the environment and avoid obstacles by changing its own shape.
Then, the experiments on leg action tracking and trajectory tracking respectively validate the precision of QuadBoat's leg action and its movements on the water surface.
Finally, the experiments on visual-feedback tracking and object pickup vefify the robot's ability to track targets using camera information and pick up objects on water surface.

	\def\demonimgwidth{0.25}
	\begin{figure*} [htpb]
		\centering
		\subfloat[]{
			\centering
			\includegraphics[width=\demonimgwidth\textwidth]{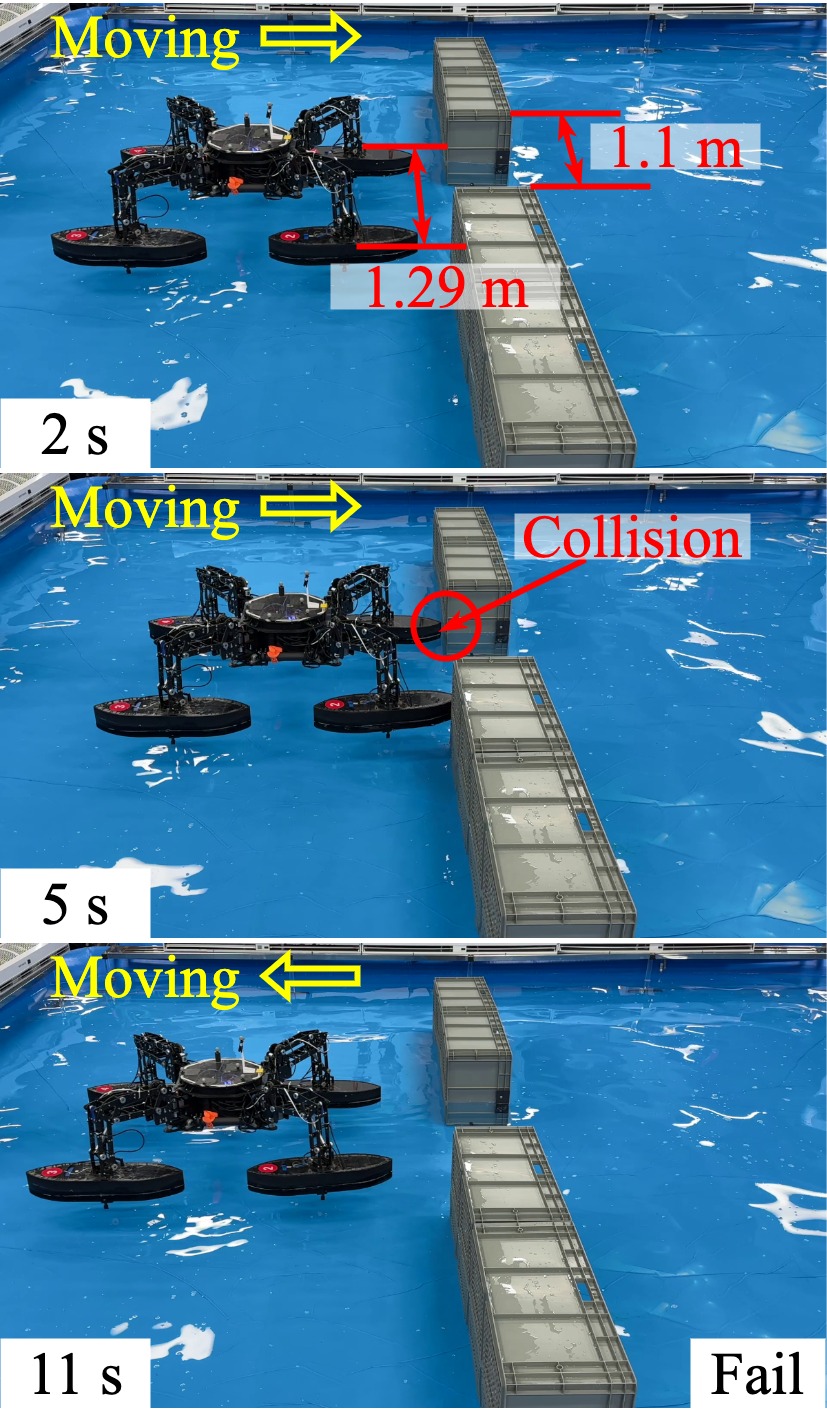}
		}
		\subfloat[]{
			\centering
			\includegraphics[width=\demonimgwidth\textwidth]{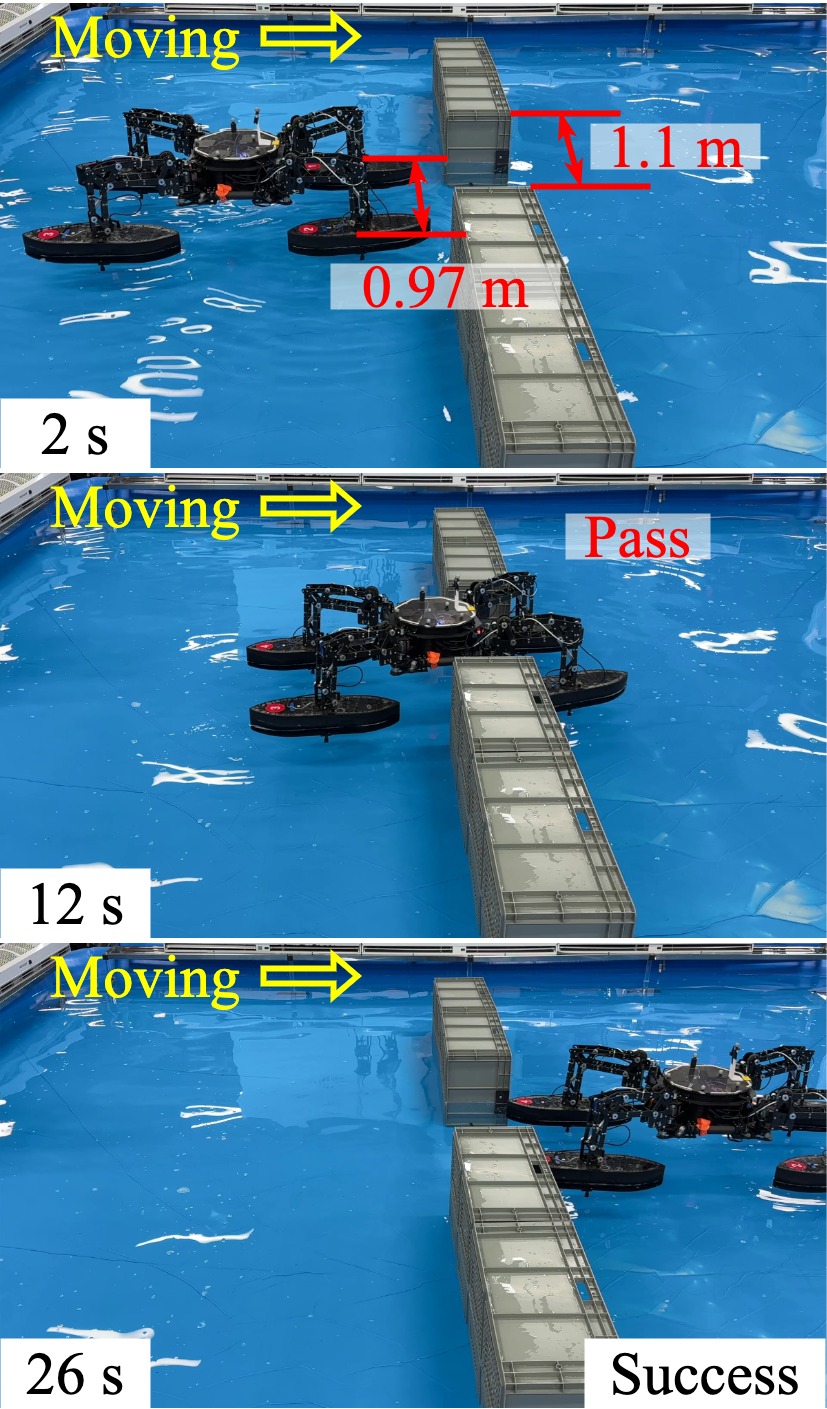}
		}
		\subfloat[]{
			\centering
			\includegraphics[width=\demonimgwidth\textwidth]{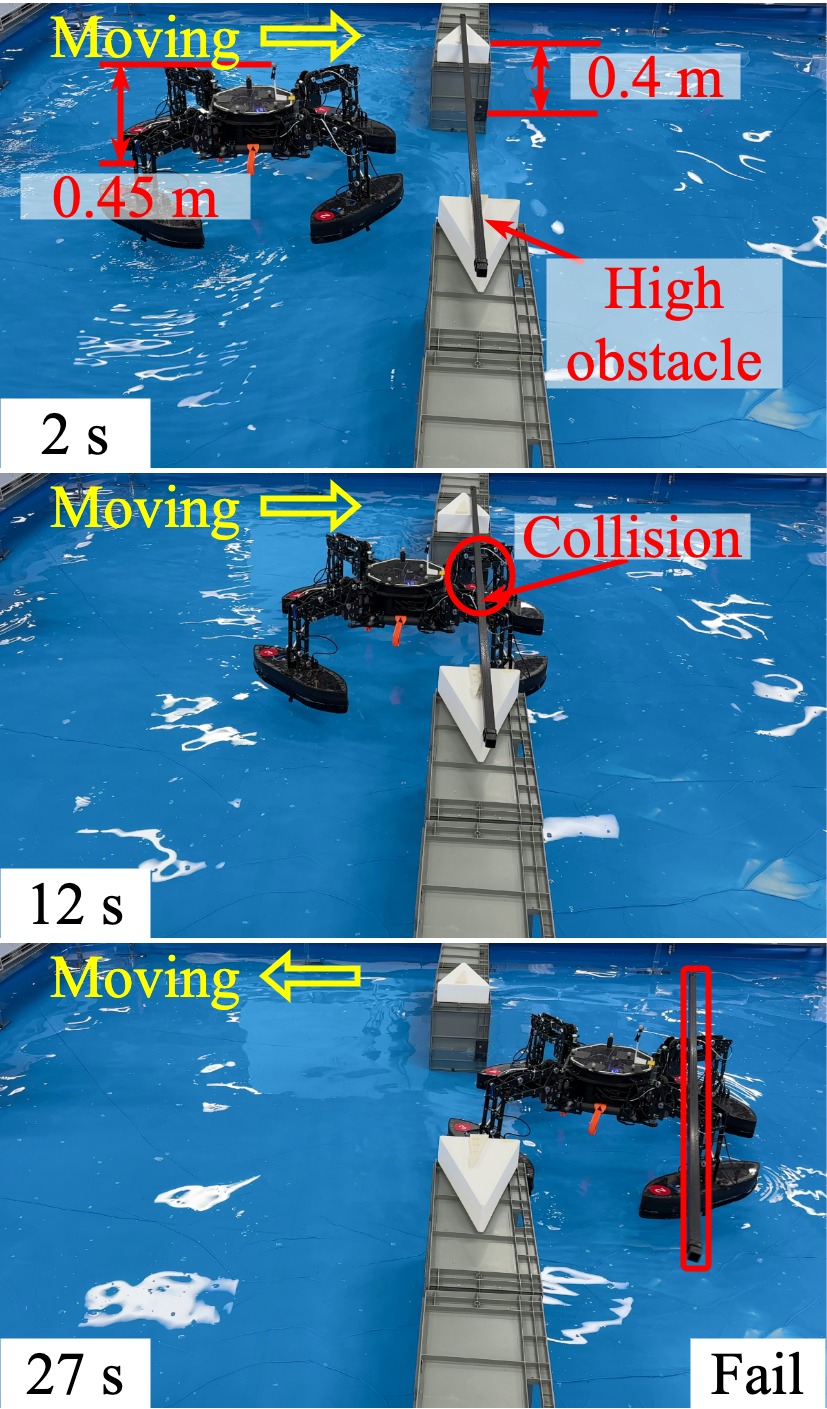}
		}
		
		\subfloat[]{
			\centering
			\includegraphics[width=\demonimgwidth\textwidth]{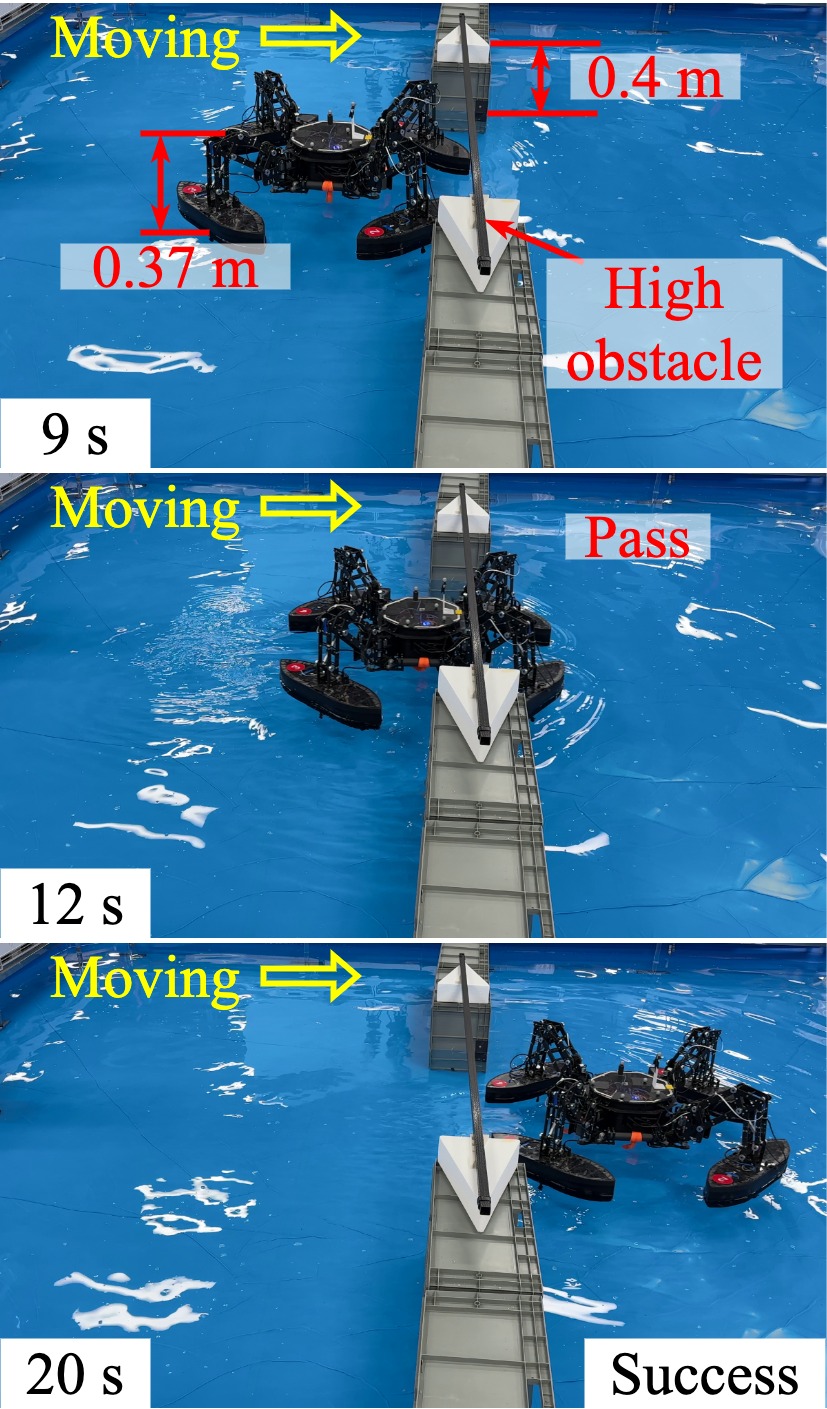}
		}
		\subfloat[]{
			\centering
			\includegraphics[width=\demonimgwidth\textwidth]{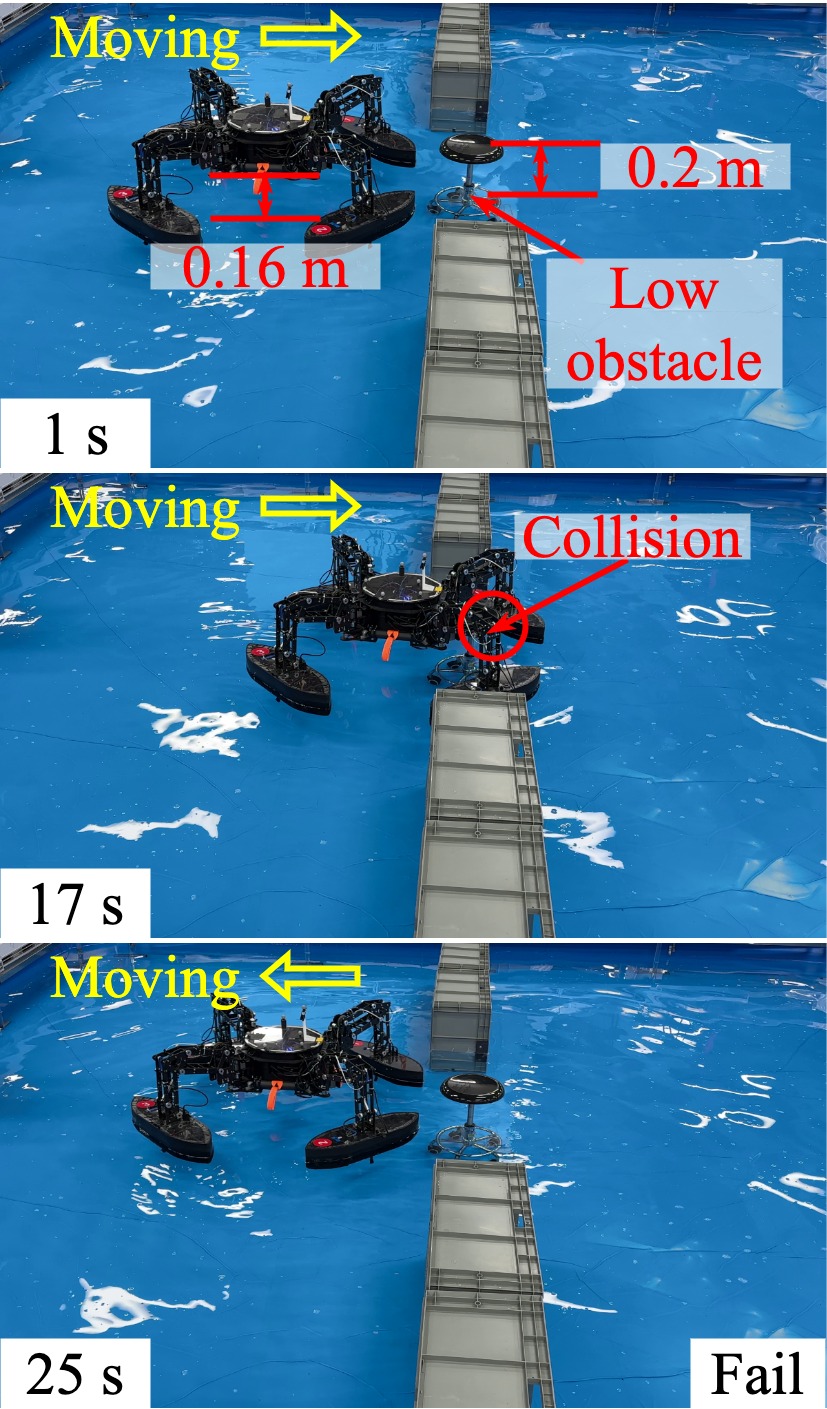}
		}
		\subfloat[]{
			\centering
			\includegraphics[width=\demonimgwidth\textwidth]{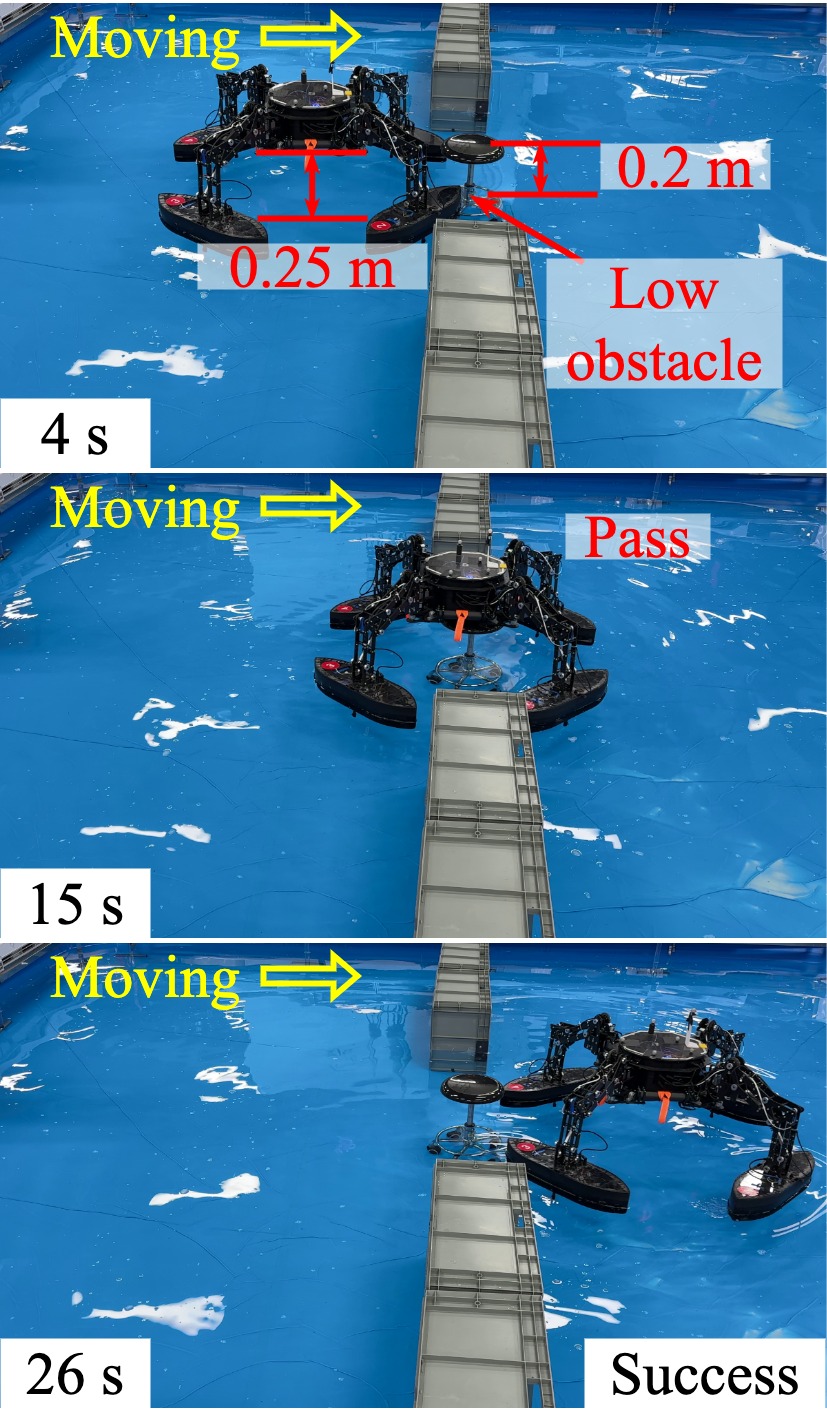}
		}
		\caption{ Experimental results of QuadBoat's feature demonstration.
			Panel (a) and (b): failed and successful attempts of QuadBoat passing through narrow passages, respectively. 
			Panel (c) and (d): failed and successful attempts of QuadBoat crawling under high obstacles, respectively.
			Panel (e) and (f): failed and successful attempts of QuadBoat crossing over low obstacles, respectively.}
		\label{fig:SystemVerificationExp}
	\end{figure*}

	\begin{sidewaystable}[htbp] 
		\tiny 
		\centering
		\caption{ERRORS IN THE TRAJECTORY TRACKING AND LEG ACTION TRACKING EXPERIMENTS}
		\label{tab:track_error}
		\renewcommand\arraystretch{1.5}
		\setlength{\tabcolsep}{1mm}
			\begin{tabular}{cc|cccccccccccc}
				\toprule[1.5pt]
				\multirow{2}{*}{\textbf{\begin{tabular}[c]{@{}c@{}}Experi\\ ment\end{tabular}}}             & \multirow{2}{*}{\textbf{Trial ID}} & \multicolumn{6}{c|}{\textbf{Trajectory tracking distance error (m)}}                                                                                                                                                                                & \multicolumn{6}{c}{\textbf{Heading angle tracking error (rad)}}                                                                                                                                                                \\
				&                                    & \textbf{Mean} & \textbf{\begin{tabular}[c]{@{}c@{}}Standard\\ deviation\end{tabular}} & \textbf{Minimum} & \textbf{Median} & \textbf{Maximum} & \multicolumn{1}{c|}{\textbf{\begin{tabular}[c]{@{}c@{}}Interquartile\\ range (Q3-Q1)\end{tabular}}} & \textbf{Mean} & \textbf{\begin{tabular}[c]{@{}c@{}}Standard\\ deviation\end{tabular}} & \textbf{Minimum} & \textbf{Median} & \textbf{Maximum} & \textbf{\begin{tabular}[c]{@{}c@{}}Interquartile\\ range (Q3-Q1)\end{tabular}} \\ \hline
				\multirow{3}{*}{\textbf{Circle}}                                                            & Circle 1                           & 0.0128         & 0.0076                                                                 & 0.0002              & 0.0125           & 0.0582            & 0.0091                                                                                               & 0.0033         & 0.0063                                                                 & -0.0523           & -0.0004             & 0.1011            & 0.004                                                                          \\
				& Circle 2                           & 0.0125         & 0.0065                                                                 & 0.0001              & 0.0131           & 0.0542             & 0.0076                                                                                               & 0.0077         & 0.0034                                                                 & -0.0332           & -0.0077          & 0.0029            & 0.0039                                                                          \\
				& Circle 3                           & 0.0128         & 0.0068                                                                 & 0.0002              & 0.0127           & 0.076            & 0.0077                                                                                               & 0.0083         & 0.0033                                                                 & -0.0337           & -0.008          & 0.0157              & 0.0028                                                                          \\ \hline
				\multirow{3}{*}{\textbf{\begin{tabular}[c]{@{}c@{}}Figure\\ -eight\end{tabular}}}        & Eight 1                            & 0.0165         & 0.0187                                                                 & 0.0004            & 0.0117           & 0.1894            & 0.0092                                                                                               & 0.0045         & 0.0101                                                                 & -0.0647            & 0.0004             & 0.1115            & 0.0047                                                                          \\
				& Eight 2                            & 0.0125         & 0.0082                                                                 & 0.0006            & 0.0102            & 0.065            & 0.0102                                                                                                & 0.0045         & 0.0095                                                                 & -0.0776            & -0.0001             & 0.0784            & 0.0052                                                                          \\
				& Eight 3                            & 0.0127         & 0.0066                                                                 & 0.0004            & 0.0111           & 0.0508            & 0.008                                                                                               & 0.0081         & 0.0053                                                                 & -0.041           & -0.0074          & 0.0253            & 0.0053                                                                          \\ \hline
				\multirow{3}{*}{\textbf{\begin{tabular}[c]{@{}c@{}}Leg \\ action \\ tracking\end{tabular}}} & $x_0 z_0$                                 & 0.0012          & 0.00071                                                                 & 0.00001            & 0.00108           & 0.00381            & 0.00102                                                                                               & \textbackslash         & \textbackslash                                                                 & \textbackslash            & \textbackslash             & \textbackslash            & \textbackslash                                                                          \\
				& $x_0 y_0$                                 & 0.00296          & 0.00125                                                                 & 0.00003           & 0.00286            & 0.00642            & 0.00192                                                                                               & \textbackslash        & \textbackslash                                                                 & \textbackslash            & \textbackslash             & \textbackslash            & \textbackslash                                                                          \\
				& $y_0 z_0$                                 & 0.00227          & 0.00133                                                                & 0.00003            & 0.00214          & 0.01233           & 0.00161                                                                                               & \textbackslash         & \textbackslash                                                                 & \textbackslash           & \textbackslash         & \textbackslash            & \textbackslash                                                                          \\ 
				\bottomrule[1.5pt]
			\end{tabular}
	\end{sidewaystable}

\subsection{Maneuverability Demonstration}
The comparison between panels (a) and (b) demonstrates QuadBoat's ability to adjust the width span of its four legs to adapt to narrow passages in the environment.
Considering the complexity of obstacles in rescue environments, this experiment explores three potential scenarios involving QuadBoat in rescue operations: navigate through narrow passages, crawling under obstacles, and crossing over obstacles. 
Fig. \ref{fig:SystemVerificationExp} illustrates the QuadBoat's capability of changing shape to overcome challenges in these scenarios.
In each scenario, we contrasted the effects before and after adjusting the posture of QuadBoat, leading to several noteworthy observations.

\subsubsection{Narrow Passage}

The comparison between panels (a) and (b) demonstrates QuadBoat's ability to adjust the width span of its four legs to adapt to narrow passages in the environment.
In the depicted scenario, there is a narrow passage, only 1.1 m wide, from the left water area to the right, which resembles real-world scenarios such as narrow waterway.
In the underactuated mode shown in panel (a), QuadBoat's body width is 1.29 m, which is greater than the passage width, making it unable to pass through.
However, in panel (b), by rotating each leg's HEE joint inward by just 15 degrees, QuadBoat can reduce its width to 0.97 m, successfully passing through.


\subsubsection{Obstacle at Heights}
Panels (c) and (d) simulate environments with tall obstacles, reminiscent of real-world scenarios such as bridges and overwater buildings.
As QuadBoat traverses the passage from left to right, it encounters a crossbar positioned 0.4m above the water surface.
In panel (c), the highest point of QuadBoat is the central body, approximately 0.45m above the water surface, causing a collision with the crossbar as it passes through, dislodging the crossbar. 
However, through leg action, QuadBoat can lower its height to 0.37m, enabling successful passage beneath the crossbar.
We can conclude that QuadBoat's ability to adjust its overall size through leg action significantly enhances its capability to overcome obstacles.

\subsubsection{Low Obstacle}
In panels (e) and (f), obstacles with a height of 0.2m above the water surface are simulated, representing real-world floating objects, rocks, etc. 
In panel (e), the lowest point of QuadBoat's central body is approximately 0.16m above the water surface, resulting in a collision with the obstacle while navigating through the passage. 
However, QuadBoat's central body can be raised to a height of 0.25m above the water surface, allowing it to successfully pass over the obstacle.
Compared to existing research on omnidirectional USVs (including some transformable USVs \cite{Zhang2022a}), QuadBoat demonstrates greater adaptability to obstacles of varying heights, thereby offering greater potential for application in rescue environments.

\begin{figure} [htpb]
	\centering
	\includegraphics[width=0.6\linewidth]{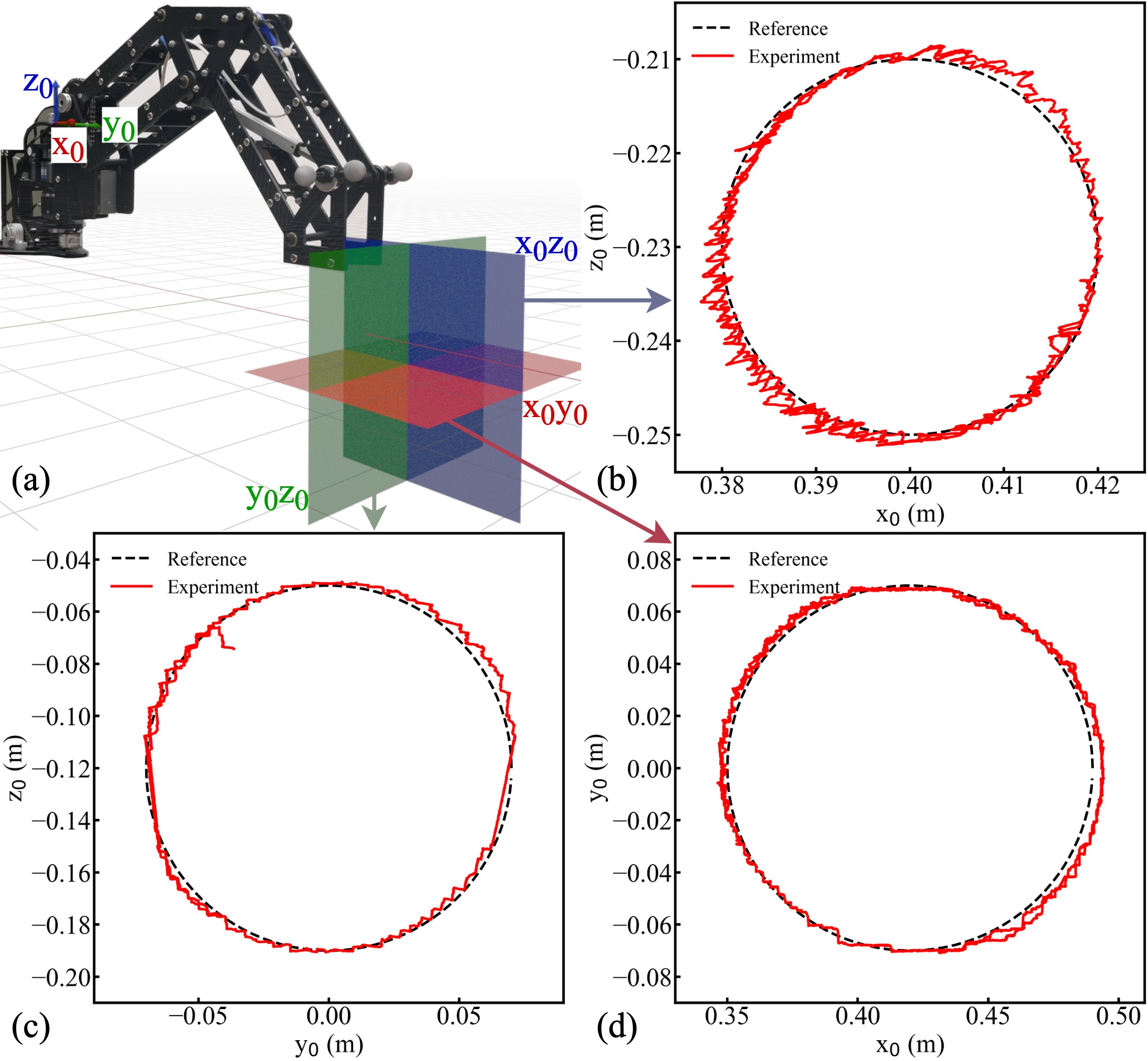}
	\caption{ Circular trajectories generated by the QuadBoat's leg actions in $x_0 z_0$, $x_0 y_0$ and $y_0 z_0$ planes.}
	\label{fig:leg_track_path}
\end{figure}

\begin{figure} [htbp] 
	\centering
	\includegraphics[width=0.6\linewidth]{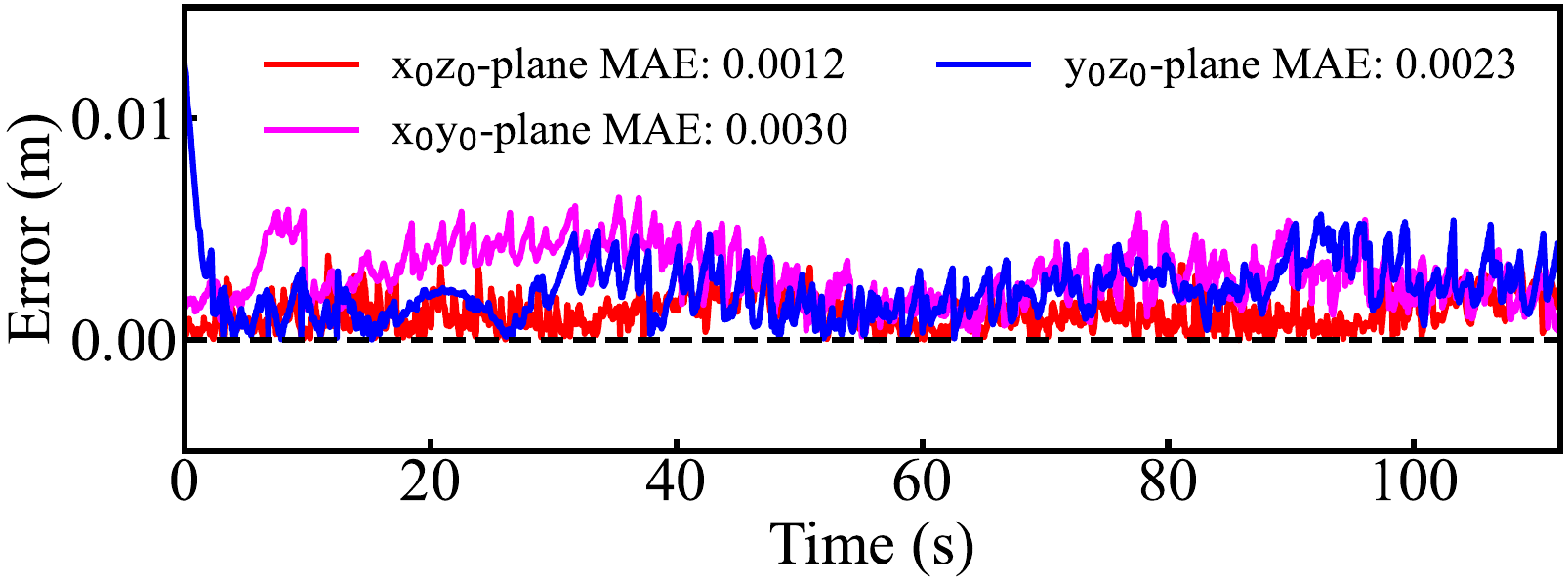}
	\caption{Trajectories tracking errors of the leg action experiment. }
	\label{fig:leg_track_error}
\end{figure}

\subsection{Leg Action Tracking}
The previous experiments require precise motion control of the four legs, and in this section we validate our controller design through trajectory tracking experiments. 
In this experiment, under the control of the aforementioned inverse kinematics-based joint controllers, the LEP can accurately track desired circular trajectories within three coordinate planes ($x_0z_0$, $x_0y_0$, $y_0z_0$). 
Due to the toroidal shape of the workspace, we choose smaller trajectory diameter in the constrained $x_0z_0$ plane, specifically setting the diameters of the three target circular trajectories as follows: $d_{x_0z_0} = 0.04 \text{m}$, $d_{x_0y_0} = 0.14 \text{m}$, $d_{y_0z_0} = 0.14 \text{m}$.
Each trial lasts for over 2 minutes.

Fig. \ref{fig:leg_track_path} illustrates the tracking results of circular trajectories in the three trials, where panel (a) depicts the coordinate system utilized in the experiment, namely the base frame $O_0-x_0 y_0 z_0$.
Since this experiment does not involve the orientation of the LEP, Fig. \ref{fig:leg_track_error} only displays the tracking distance errors, while detailed statistical distribution data is provided in Table \ref{tab:track_error}. 
The minimum and maximum tracking distance errors for the three trials are as follows: 0.01 mm and 3.81 mm for the $x_0z_0$ plane trial, 0.03 mm and 6.42 mm for the $x_0y_0$ plane trial, and 0.03 mm and 12.33 mm for the $y_0z_0$ plane trial. 
In the initial stages, the errors are relatively large due to the experimental path not yet converging near the reference trajectory points. 
Considering the different trajectory diameter for each trial, besides comparing the maximum and minimum errors, we also utilize the interquartile range (Q3?Q1, IQR) and standard deviation (SD) of the errors as evaluation criteria.
The IQR of these three trials are 1.02 mm, 1.92 mm, and 1.61 mm, respectively.

We compared the tracking errors of the three trials and reached the following conclusions.
First, compared to the range of the workspace and the size of the structures, all tracking errors are relatively small, indicating consistent performance of the legs in different directions, validating the rationality of our design.
Second, the mean errors (MAEs) in the $x_0y_0$ plane and $y_0z_0$ plane are 2.96 mm (SD 1.25mm) and 2.27 mm (SD 1.33 mm), respectively, both significantly larger than the error in the $x_0z_0$ plane (MAE 1.2mm, SD 0.71mm). 
This difference is not readily apparent in the path plot (Fig. \ref{fig:leg_track_path}). 
The reason for this is that the leg action in the former two planes requires the use of the HEE joint, which is the furthest from the LEP. 
Errors at the HEE joint are amplified considerably after transmission.

\def\trajtrackwidth{0.32}
\begin{figure}[htbp]      
	\centering
	\begin{minipage}[b]{\trajtrackwidth\linewidth}
		\centering
		\subfloat[]{
			\includegraphics[width=1\linewidth]{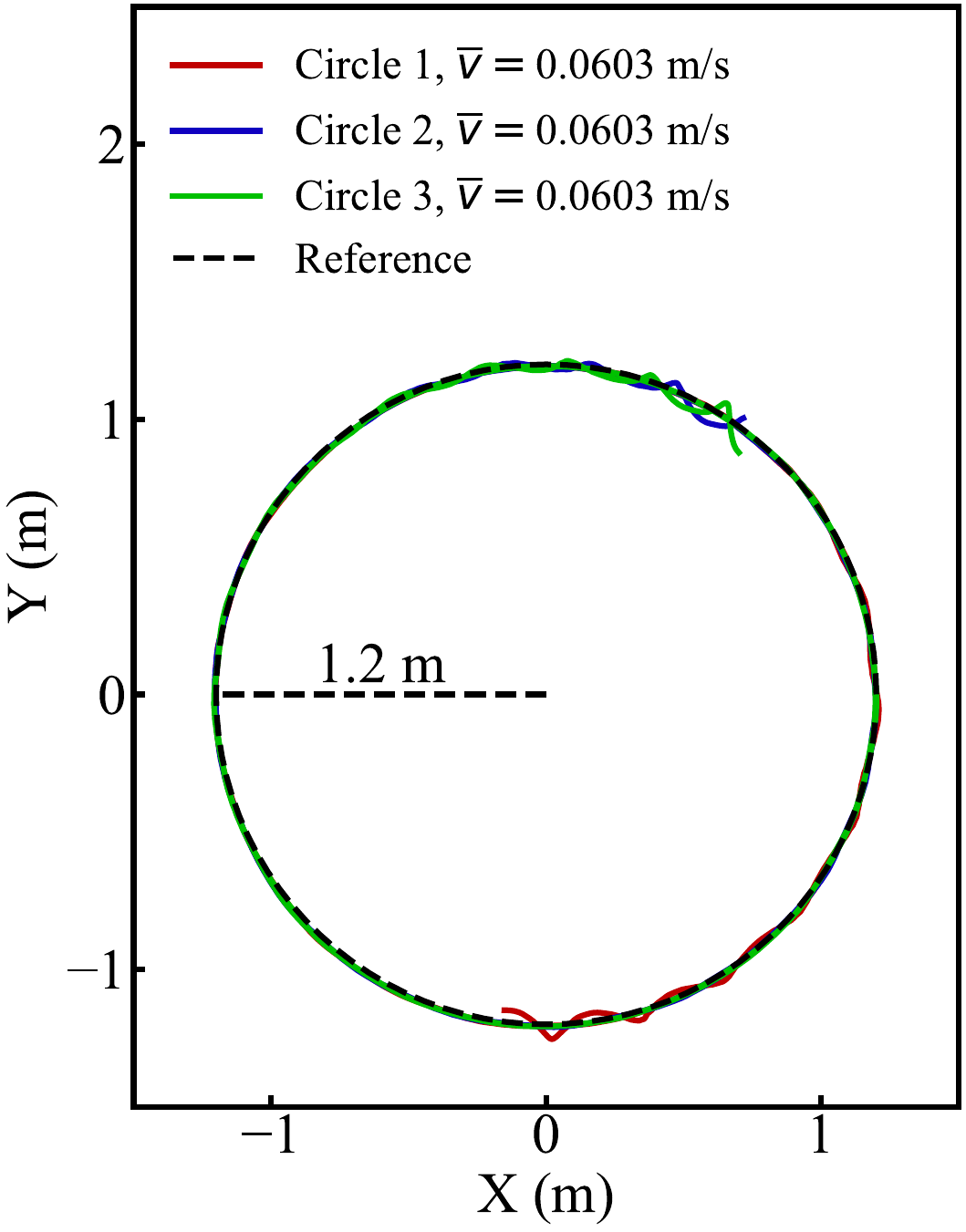}
		}
	\end{minipage}
	\begin{minipage}[b]{\trajtrackwidth\linewidth}
		\centering
		\subfloat[]{
			\includegraphics[width=1\linewidth]{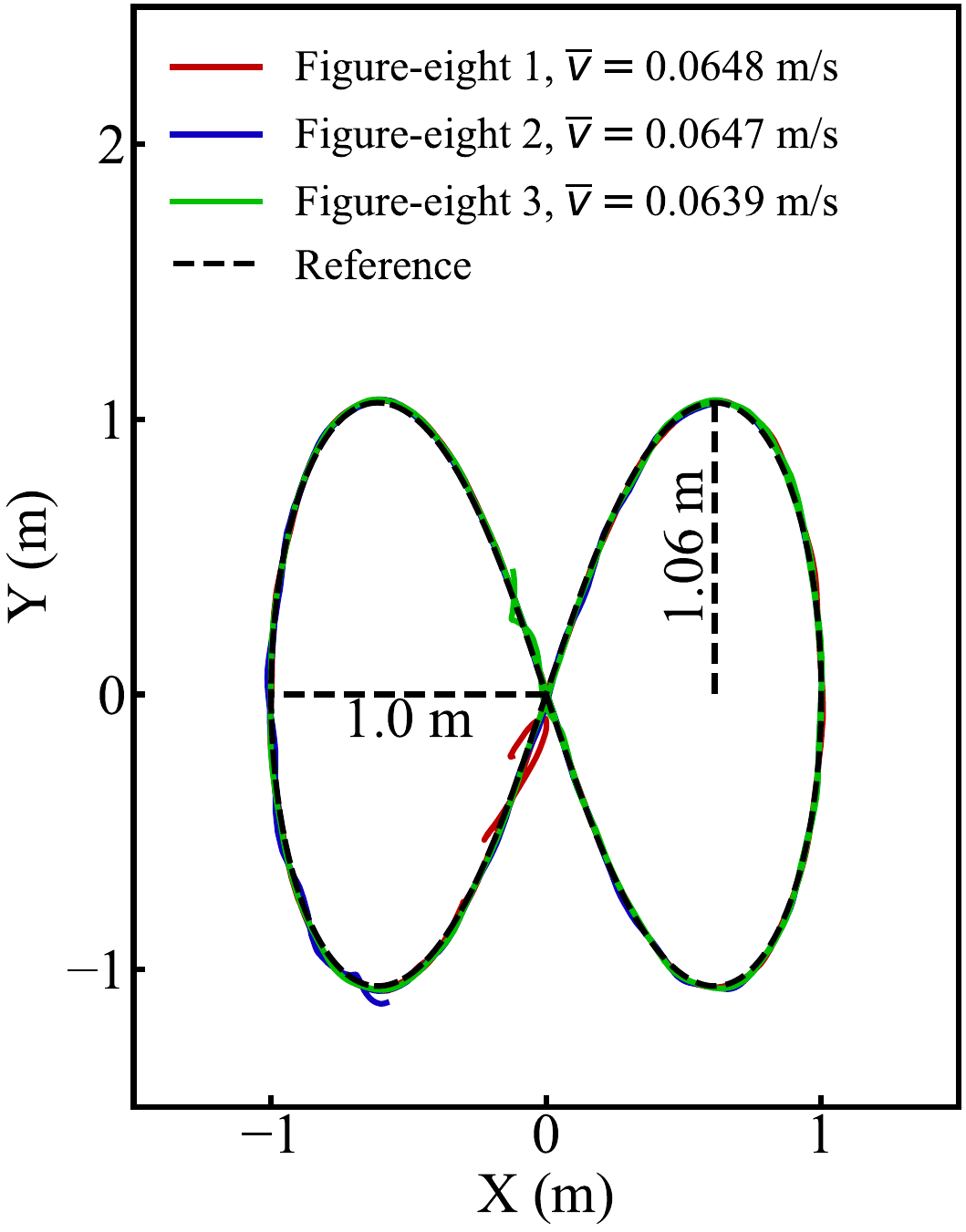}
		}
	\end{minipage}
	
	\begin{minipage}[b]{\trajtrackwidth\linewidth}
		\centering
		\subfloat[]{
			\includegraphics[width=1\textwidth]{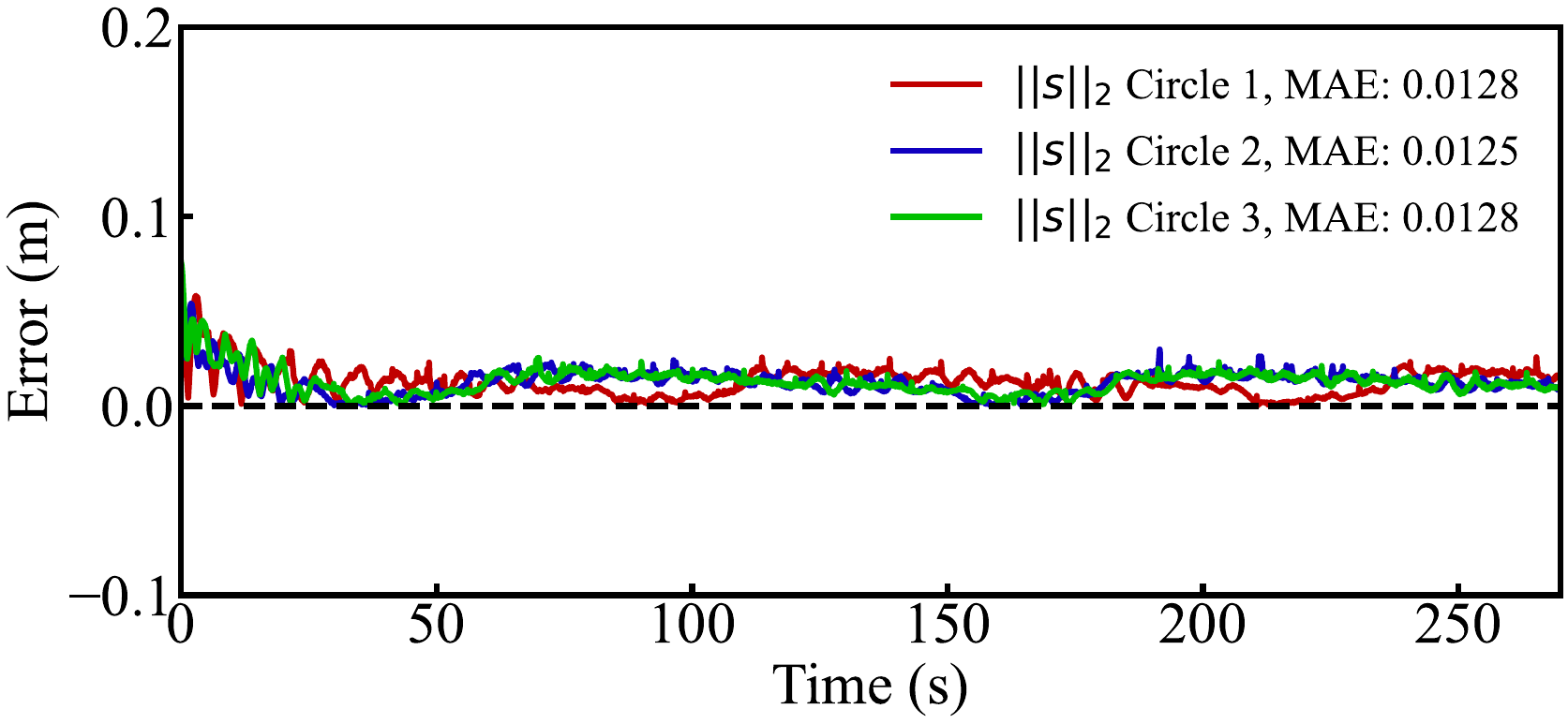}
		}
	\end{minipage}
	\begin{minipage}[b]{\trajtrackwidth\linewidth}
		\subfloat[]{
			\includegraphics[width=1\textwidth]{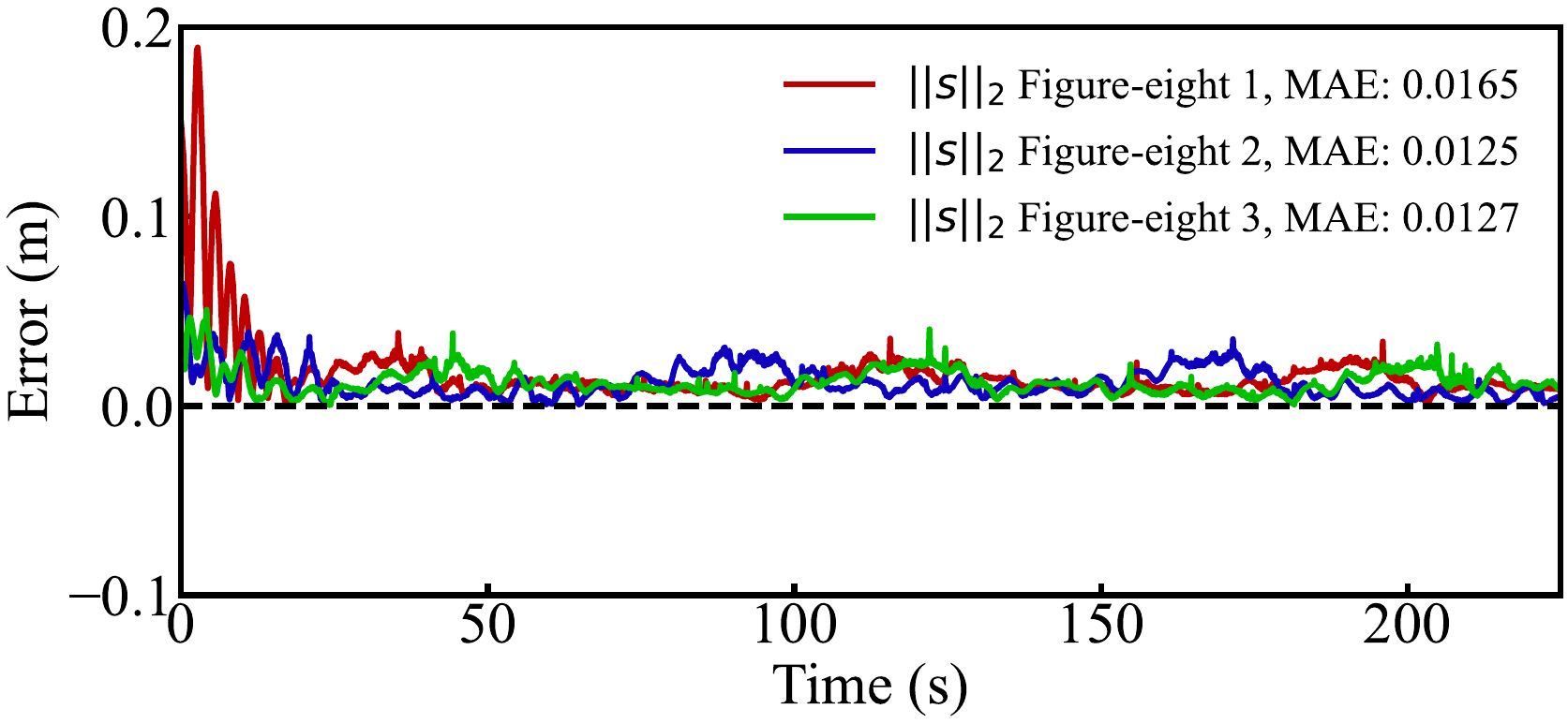}
		}
	\end{minipage} 
	
	\begin{minipage}[b]{\trajtrackwidth\linewidth}
		\centering
		\subfloat[]{
			\includegraphics[width=1\textwidth]{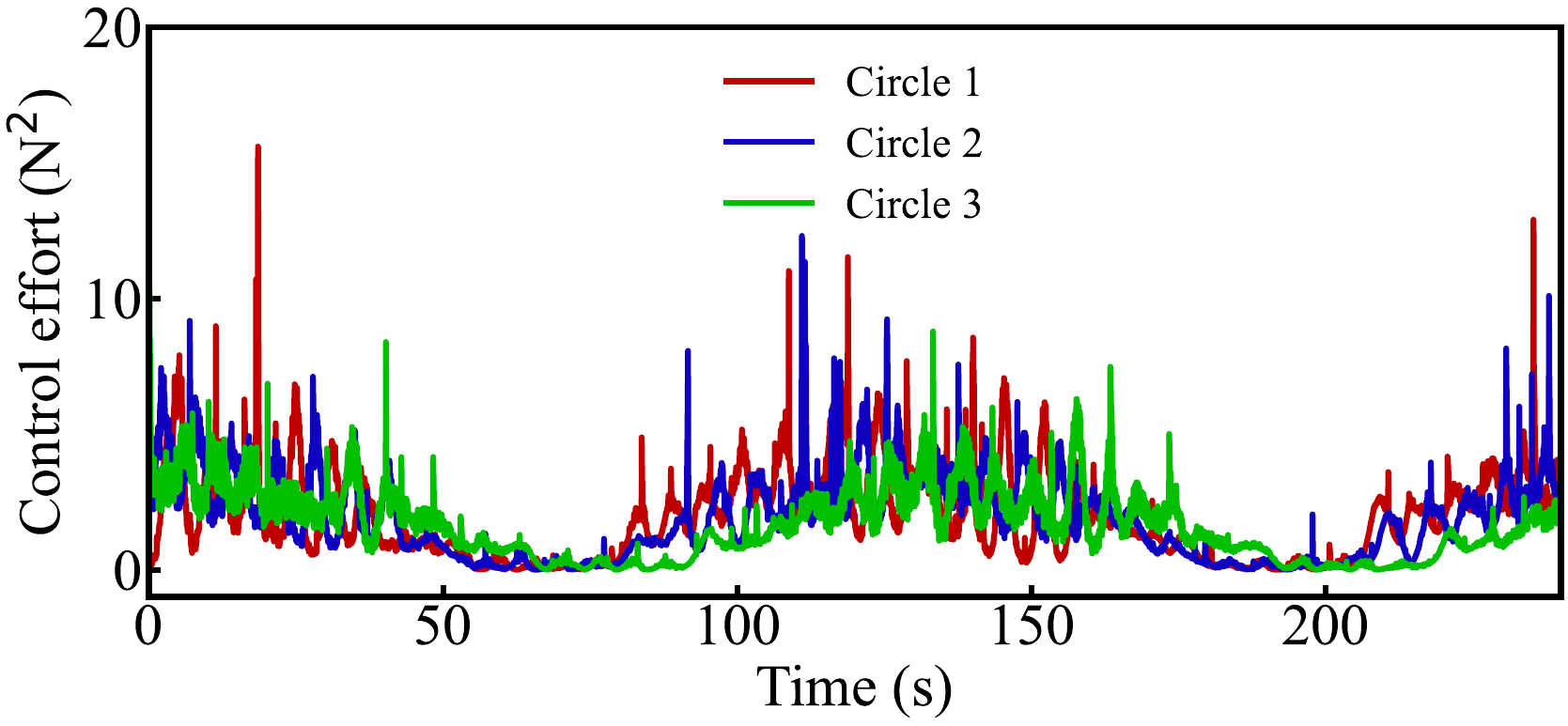}
		}
	\end{minipage}
	\begin{minipage}[b]{\trajtrackwidth\linewidth}
		\subfloat[]{
			\includegraphics[width=1\textwidth]{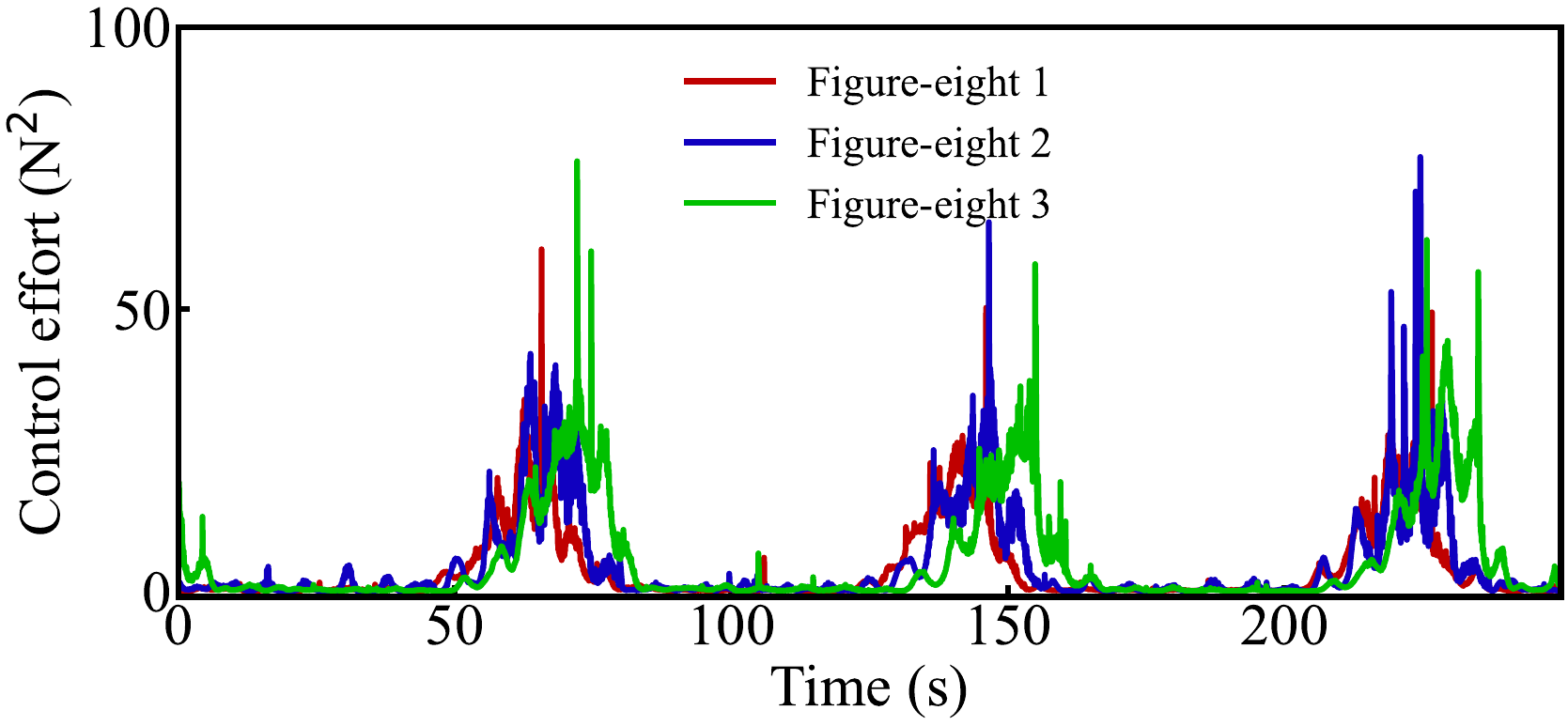}
		}
	\end{minipage}   
	\caption{Experimental paths, errors and control efforts in tracking  (a)(c)(e) the circular and (b)(d)(f) the eight-shape trajectories.
		The average velocities $\bar{v}$ and the mean absolute errors (MAEs) are calculated. }
\label{fig:tracking}  
\end{figure}

\subsection{Trajectory Tracking}
To validate our designed cascade MPC-PID controller for QuadBoat's overall movement, we design two trajectories in this experiment: circular and figure-eight trajectories.
In the $6\ m \times 6\ m$ experimental pool, the diameter of the circular trajectory is 2.4 m, and the figure-eight trajectory has a length of 2 m along the x-axis and 2.12 m along the y-axis.
Each trajectory is repeated three times, resulting in a total of 6 trials, with each trial lasting over 6 minutes.

Fig. \ref{fig:tracking} presents the experimental results of trajectory tracking, including the experimental trajectories, distance errors, and control efforts, with corresponding statistical data provided in Table \ref{tab:track_error}. 
It is worth noting that our focus here is on achieving the best tracking performance rather than high sailing speed. Hence, a tracking speed of approximately 0.06 m/s was set. In practice, the Quadboat can move much faster, reaching up to 0.8 m/s, as shown in Table \ref{tab:spec}.
The results indicate that the errors for all trials are small, with minimum and maximum errors as follows: 0.1 mm and 76 mm for the circular trajectory, and 0.4mm and 189.4mm for the figure-eight trajectory.
Similar to the previous experiment, we utilized the IQR and SD as evaluation criteria. 
The IQR for the circular trajectory ranges from 7.7mm to 9.1mm, while for the figure-eight trajectory, it ranges from 8mm to 10.2mm.

By comparing the experimental results of different trials, we can draw the following conclusions.
First, compared to other experiments with omnidirectional unmanned boats, our tracking performance is excellent, verifying the design.
We can contrast our circular trajectory tracking experiments with those in \cite{Zhang2022a}, where the MAE is reported as (68~75.6 mm), and in \cite{Xue2023}, the errors are (distance MAE: -34-41 mm with SD 86-173 mm) and (orientation MAE: 0.002-0.029 radians with SD 0.065-0.132 radians). In comparison, our distance error ranges from 12.5-12.8 mm with a SD of 6.5-7.6 mm, and orientation error ranges from 0.0033-0.0083 radians with a SD of 0.0033-0.0063 radians. 
Both of our errors are smaller, indicating better performance. 
Second, the maximum MAE for the figure-eight trajectory is slightly larger than that for the circular trajectory (16.5 mm vs. 12.8 mm), and the maximum control effort (about 80 $N^2$ vs. about 16 $N^2$ ) is also higher, attributed to the complexity of the former trajectory.
Third, there are three peaks in the control effort of the figure-eight trials (panel (f)), corresponding to the turns in the trajectory.

\begin{figure} [htpb]
\centering
\subfloat[]{
	\centering
	\includegraphics[width=0.45\linewidth]{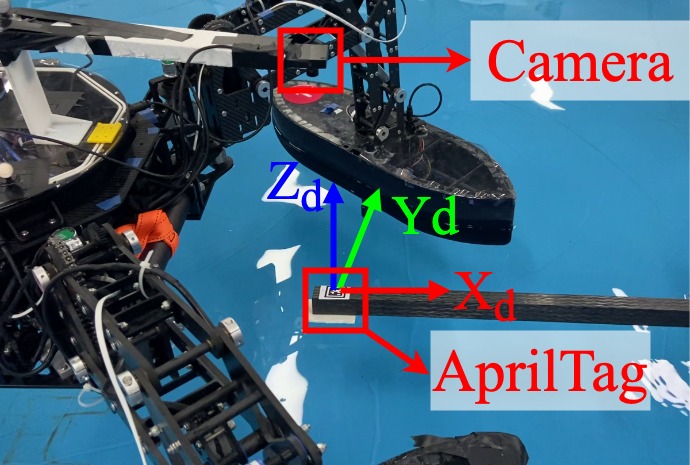}
}
\subfloat[]{
	\centering
	\includegraphics[width=0.45\linewidth]{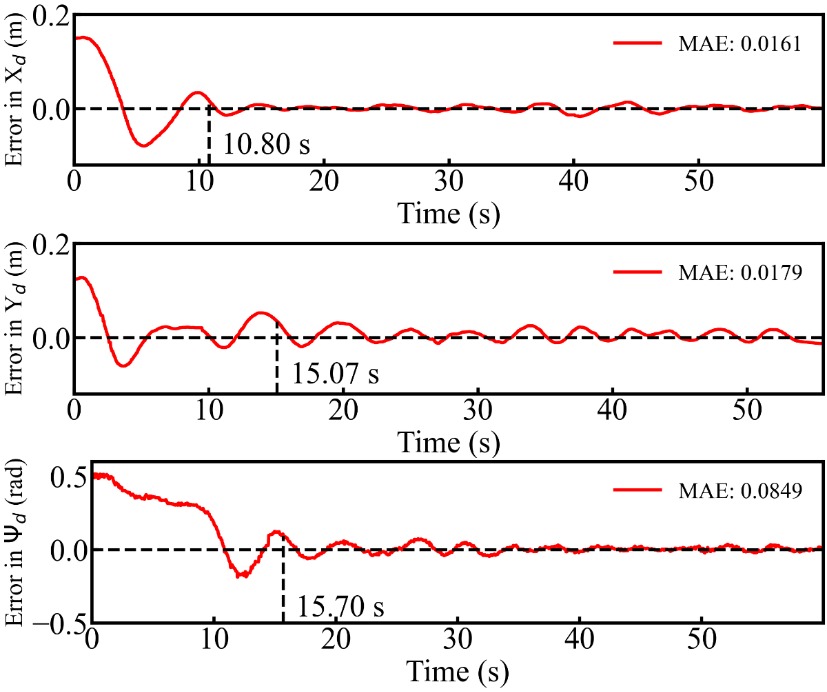}
}
\caption{ (a) Experimental scene of the visual feedback tracking. (b) Tracking errors along the $X_d$, $Y_d$ and $\Psi_d$ (yaw) axes.}
\label{fig:camera_track_path}
\end{figure}

\subsection{Visual Feedback Tracking}
When QuadBoat performs rescue missions, it identifies drowning victims through cameras and subsequently tracks and picks up the targets.
To verify QuadBoat's ability to track water surface targets under visual guidance, we set up a long pole on the water surface with an apriltag fixed at the end for QuadBoat to recognize and track. 
Fig. \ref{fig:camera_track_path} (a) depicts the experimental setup, including QuadBoat, its camera, the apriltag representing the target point, and the coordinate system established on it. 
In this experiment, QuadBoat's control objective is to align the center of the camera's field of view with the center of the target apriltag. 
Building on the foundation of the aforementioned movement trajectory tracking experiment, we introduce initial position or orientation deviations in the x, y, and yaw directions to induce a step response in the system. 
The initial deviations are approximately 0.15 m, 0.125 m, and 0.49 radians, respectively.

Fig. \ref{fig:camera_track_path} (b) presents the results of the step responses in the three directions, from which we draw the following conclusions.
First, the QuadBoat system based on visual feedback demonstrates excellent capability in tracking water surface targets. Specifically, the tracking MAE in the three directions are 16.1 mm, 17.9 mm, and 0.0849 radians, respectively, all significantly smaller than the body length of QuadBoat ($>$ 1188 mm). 
The errors rapidly converge to 0 and stabilize near 0. Second, considering the settling times of 10.8 s, 15.07 s, and 15.70 s in the respective directions, it is advisable for QuadBoat to wait at least 16 seconds before executing the object pickup action.

\def\pickimgwidth{0.32}
\begin{figure} [htpb]
\centering
\subfloat[]{
	\centering
	\includegraphics[width=\pickimgwidth\linewidth]{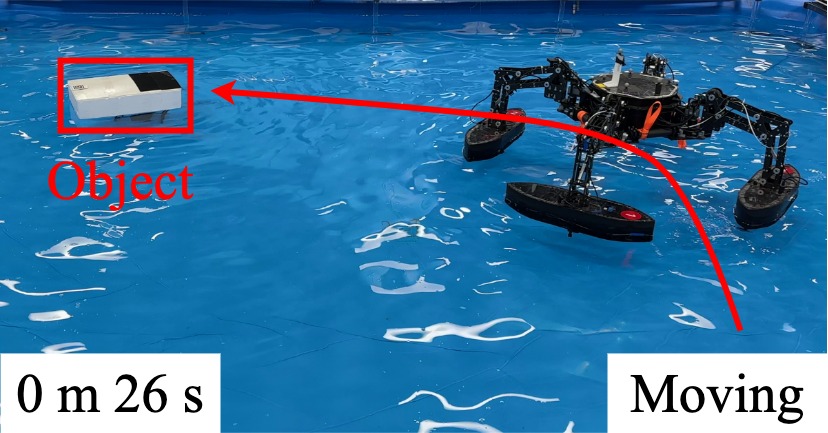}
}
\subfloat[]{
	\centering
	\includegraphics[width=\pickimgwidth\linewidth]{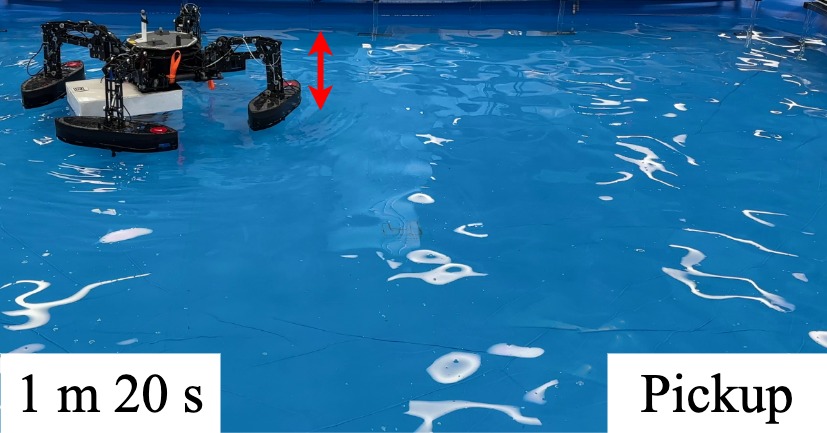}
}
\subfloat[]{
	\centering
	\includegraphics[width=\pickimgwidth\linewidth]{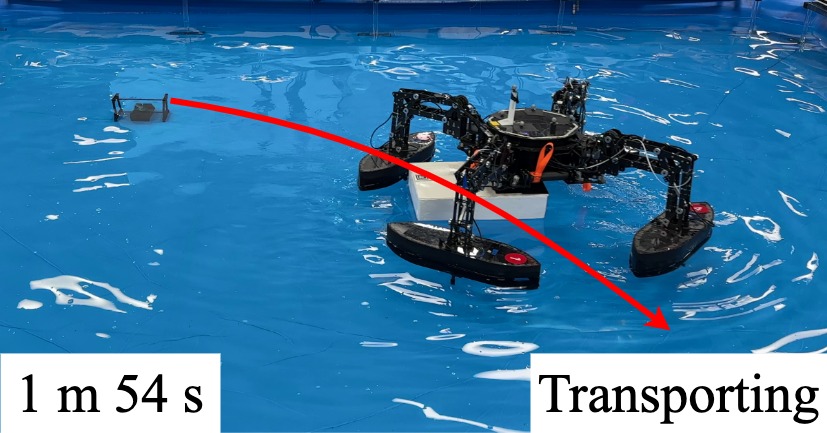}
}
\newline
\subfloat[]{
	\centering
	\includegraphics[width=\pickimgwidth\linewidth]{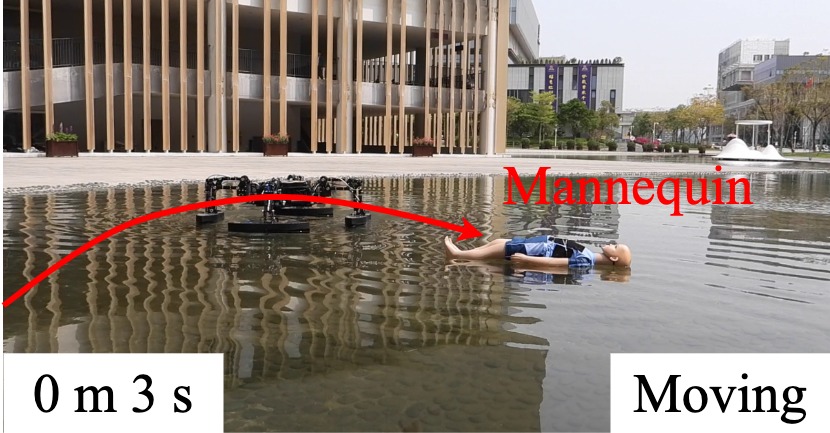}
}
\subfloat[]{
	\centering
	\includegraphics[width=\pickimgwidth\linewidth]{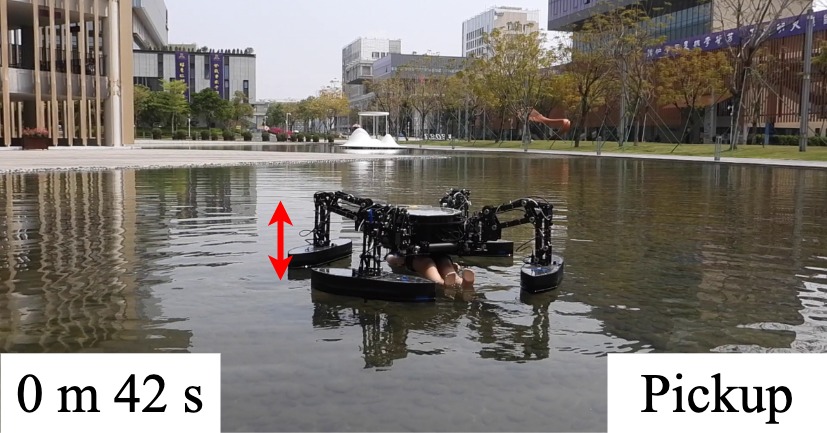}
}
\subfloat[]{
	\centering
	\includegraphics[width=\pickimgwidth\linewidth]{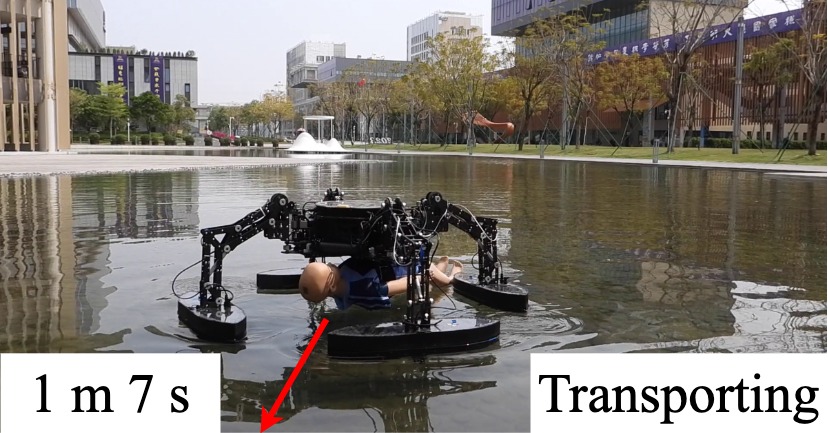}
}
\caption{ Indoor (a) (b) (c) and outdoor (d) (e) (f) experimental processes of object pickup conducted by the QuadBoat.}
\label{fig:pickup}
\end{figure}

\subsection{Object Pickup}
Based on the validated system design and control methods, in this experiment, we controlled QuadBoat to track targets on the water surface and lift them out of the water to achieve water surface rescue.
The experiments were conducted both indoors and outdoors, with each experiment repeated 10 times, and success rates and average completion times for the successful trials were recorded. 
The target for indoor experiments was foam blocks, while for outdoor experiments, a mannequin model was used, assisted by apriltags to aid camera recognition.

The process for both experiments includes three steps: moving toward and capturing the target, pickup, and transporting the target to the shore, as shown in Fig. \ref{fig:pickup}.
Moving toward the target and transportation are remotely controlled from the shore, while tracking and lifting based on visual feedback are automatic. 
The statistical results of the experiments are as follows: the success rate is $70 \%$ for indoor trials and $50 \%$ for outdoor trials, with average completion times of 116 s for indoor trials and 132 s for outdoor trials.
From these results, we can conclude: First, QuadBoat is capable of accurately and quickly lifting target victims from the water surface to the shore, thus demonstrating its suitability for rescue operations. 
Second, the success rate and completion time of outdoor trials are lower and longer, respectively, compared to indoor trials, attributed to environmental factors such as wind, waves, and sunlight reflection.


\section{Conclusion}
\label{sect:conclusion}
In water rescue missions, timely removal of victims from the water is crucial for successful rescue operations.
This paper presents a bio-inspired USV, named QuadBoat, along with its leg action controller and overall movement controller based on cascaded MPC and PID controllers.
Equipped with four thrusters and the ability to adjust the posture of its four legs, QuadBoat exhibits strong maneuverability on the water surface, with both fully and under-actuated motion modes. 
Utilizing onboard cameras, QuadBoat can track and extract objects from the water, thereby executing rescue missions.
A series of experiments, including leg action tracking, trajectory tracking, and visual-based tracking, validate the high motion accuracy and target tracking capabilities of the QuadBoat system. 
Maneuverability demonstrations and object pickup experiments confirm QuadBoat's high agility and demonstrate its effectiveness in executing rescues.


In the future, it is imperative to develop QuadBoat's amphibious capabilities for transitioning from water to land. 
Additionally, research on high-speed motion control of QuadBoat is essential to meet the demands of emergency SAR scenerios. 
Furthermore, enhancing QuadBoat's visual recognition capabilities to identify victims on the water surface will be a key focus of future research.

\backmatter

\bibliography{USV}

\end{document}